\documentclass[10pt,journal,compsoc]{IEEEtran}
\usepackage{amssymb}
\usepackage{makecell}
\usepackage{graphicx}
\usepackage{stfloats}
\usepackage{algorithm}
\usepackage{algorithmic}
\usepackage{amsmath}
\usepackage{subfigure}
\usepackage{epstopdf}
\usepackage{bm}
\usepackage{multirow}
\usepackage{booktabs}  
\usepackage{threeparttable}  
\usepackage{makecell}
\usepackage{graphicx}
\usepackage{stfloats}
\usepackage{subfigure}
\usepackage{epstopdf}
\usepackage{float}
\usepackage{bm}
\usepackage{hhline}
\usepackage{fancyhdr}
\usepackage{mathtools}
\usepackage{cuted}
\usepackage{capt-of}
\usepackage{color}
\usepackage{cite} 
\ifCLASSINFOpdf
\else
\fi
\begin{document}
\title{Visual Perturbation-aware Collaborative Learning for Overcoming the Language Prior Problem}

\author{Yudong~Han,
        Liqiang Nie,~\IEEEmembership{Senior Member,~IEEE, }
        Jianhua~Yin,~\IEEEmembership{Member,~IEEE, }
        \\
        Jianlong Wu,~\IEEEmembership{Member,~IEEE, }% <-this % stops a space
        Yan Yan,~\IEEEmembership{Member,~IEEE,}% <-this % stops a space
\IEEEcompsocitemizethanks{\IEEEcompsocthanksitem Yudong Han, Jianhua Yin, and Jianlong Wu are with the School of Computer Science and Technology, Shandong University, Qingdao, 266237, China. (e-mail: hanyudong.sdu@gmail.com; jhyin@sdu.edu.cn; jlwu1992@sdu.edu.cn)
\IEEEcompsocthanksitem Liqiang Nie is a professor with Harbin Institute of Technology (Shenzhen). (e-mail: nieliqiang@gmail.com)
\IEEEcompsocthanksitem Yan Yan is currently a Gladwin Development Chair Assistant Professor in the Department of Computer Science at Illinois Institute of Technology. (e-mail: yyan34@iit.edu)
}

}
% \thanks{Manuscript received April 19, 2005; revised August 26, 2015.}


\markboth{IEEE Transactions on pattern analysis and machine intelligence, 2022}%
{Shell \MakeLowercase{\textit{et al.}}: Bare Demo of IEEEtran.cls for Computer Society Journals}



\IEEEtitleabstractindextext{%
\begin{abstract}
Several studies have recently pointed that existing Visual Question Answering (VQA) models heavily suffer from the language prior problem, which refers to capturing superficial statistical correlations between the question type and the answer whereas ignoring the image contents. Numerous efforts have been dedicated to strengthen the image dependency by creating the delicate models or introducing the extra visual annotations. However, these methods cannot sufficiently explore how the visual cues explicitly affect the learned answer representation, which is vital for language reliance alleviation. Moreover, they generally emphasize the class-level discrimination of the learned answer representation, which overlooks the more fine-grained instance-level patterns and demands further optimization. In this paper, we propose a novel collaborative learning scheme from the viewpoint of visual perturbation calibration, which can better investigate the fine-grained visual effects and mitigate the language prior problem by learning the instance-level characteristics. Specifically, we devise a visual controller to construct two sorts of curated images with different perturbation extents, based on which the collaborative learning of intra-instance invariance and inter-instance discrimination is implemented by two well-designed discriminators. Besides, we implement the information bottleneck modulator on latent space for further bias alleviation and representation calibration. We impose our visual perturbation-aware framework to three orthodox baselines and the experimental results on two diagnostic VQA-CP benchmark datasets evidently demonstrate its effectiveness. In addition, we also justify its robustness on the balanced VQA benchmark. 
% We have released the code at: \color{magenta}{\url{https://github.com/Einstone-rose/LUNA}}.
\end{abstract}

}

\maketitle

\IEEEdisplaynontitleabstractindextext

% \fi
\IEEEpeerreviewmaketitle

\section{Introduction}

\begin{figure}
    \centering
    \includegraphics[width=0.48\textwidth]{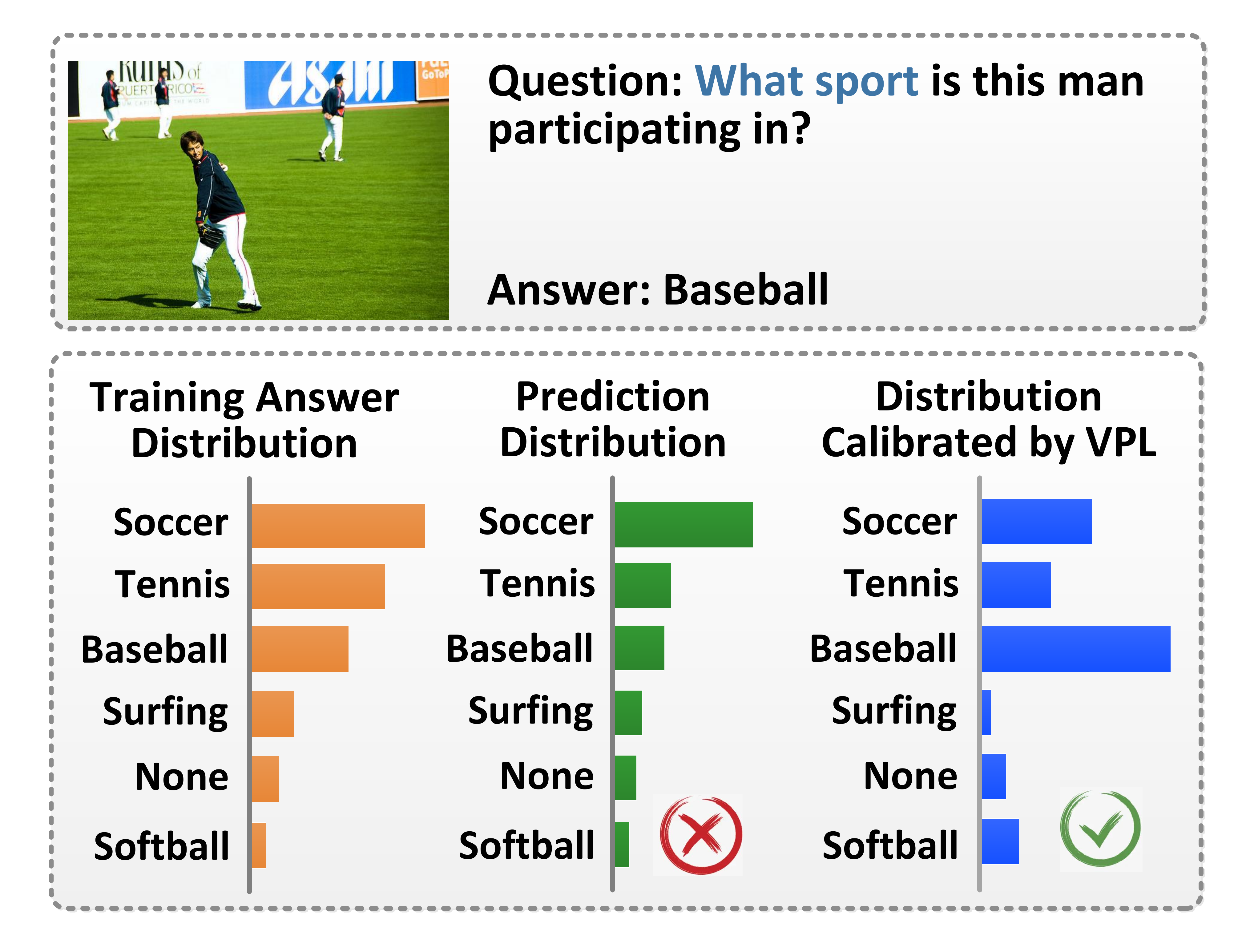}
    \vspace{-1ex}
    \caption{ Illustration of why the language prior problem arises and our solution. The answer distribution (orange bars) on the \textit{what sport} question type in the training set is biased, which makes the answer prediction (green bars) to drift towards frequent answers \textit{Soccer}. Our visual perturbation learning (VPL) leverages the fine-grained instance-specific visual clues to help calibrate the answer distribution (blue bars).}
    \label{fig:introduction}
    \vspace{-1em}
\end{figure}

\IEEEPARstart{D}{eemed} as an AI-complete task, Visual Question Answering (VQA) has become an emerging interdisciplinary research task over the past few years. It targets at automatically answering natural language questions given a visual scene. With the increasing prosperity of both computer vision and natural language processing communities, several large-scale benchmarks~\cite{DBLP:conf/iccv/AntolALMBZP15, DBLP:conf/cvpr/YangHGDS16} have been fabricated to facilitate the advancement of VQA realm, followed by a series of well-designed multi-modal fusion models~\cite{DBLP:conf/iccv/AntolALMBZP15, DBLP:conf/cvpr/YangHGDS16, DBLP:conf/cvpr/00010BT0GZ18, hu2021video, hu2021coarse}. However, some researchers have demonstrated that existing VQA models suffer from the superficial shortcut between the question (i.e., question type) and the answers, as illustrated in Fig.~\ref{fig:introduction}. As a result, the visual effect is largely overwhelmed, making the visual reasoning in VQA degenerate to a sheer language matching problem.

\begin{figure*}
    \centering
    \includegraphics[width=0.96\textwidth]{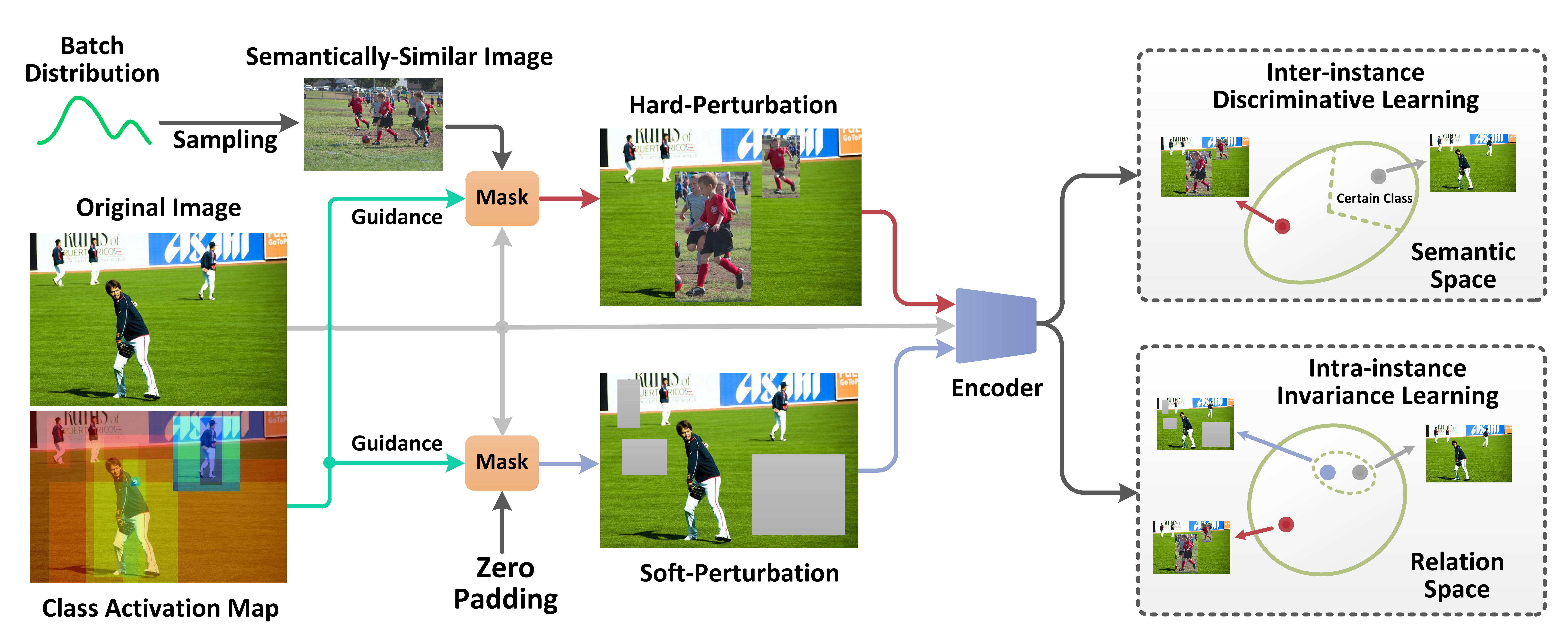}
    \caption{Illustration of visual perturbation-aware learning. With the guidance of class activation map (i.e., learned attention), two sorts of curated image features with different perturbation extents (i.e., soft perturbation and hard perturbation) are dynamically constructed. Based on this, the intra-instance invariant and inter-instance discriminative learning are jointly performed to explore the instance-level fine-grained visual characteristics and language prior problem alleviation. For the ease of illustration, we omit the question branch.}
    \label{fig:vpl}
    \vspace{-1em}
\end{figure*}

The mainstreaming methods~\cite{DBLP:conf/nips/RamakrishnanAL18, DBLP:conf/emnlp/LiangJHZ20, DBLP:conf/iccv/SelvarajuLSJGHB19, DBLP:conf/nips/WuM19, DBLP:conf/cvpr/0016YXZPZ20, DBLP:conf/aaai/PatroAN20, DBLP:conf/iccv/SelvarajuCDVPB17, DBLP:conf/ijcai/ZhuMLZWZ20} in VQA attempt to mitigate the language prior problem via boosting the image dependency, which can be essentially categorized into two groups: visual annotation-based and visual annotation-free methods. The former ones explicitly leverage external visual annotations to guide the learning of visual glimpse. For instance, HINT~\cite{DBLP:conf/iccv/SelvarajuLSJGHB19} and SCR~\cite{DBLP:conf/nips/WuM19} learn the high-caliber visual clues by optimizing the alignment between external human attention map~\cite{DBLP:conf/cvpr/ParkHARSDR18, DBLP:journals/cviu/DasAZPB17} and gradient-based importance of image regions. Despite their promising improvement, collecting such human annotations is expensive and time-consuming, which inevitably limits their adaptation across datasets. On the contrary, the visual annotation-free methods have revolutionized the aforementioned annotation-based paradigm. The popular solutions\cite{DBLP:conf/cvpr/0016YXZPZ20, DBLP:conf/iccv/SelvarajuCDVPB17} are to deploy a subsidiary question-only branch to modulate the question-image branch, which intentionally captures the language priors and provokes the potential of visual effects. Orthogonal to these two-branch studies, some researchers take full advantage of the surrogate visual clues derived from Grad-CAM~\cite{DBLP:conf/iccv/SelvarajuCDVPB17} results by utilizing the contrastive or self-supervised techniques for better visual calibration. Moreover, some data augmentation strategies are also adopted in some work~\cite{DBLP:conf/ijcai/ZhuMLZWZ20, DBLP:conf/iccv/HanWSH021} to adjust the image distribution and mitigate the imbalance partial to language prior. 

Despite their exciting prospects of alleviating the language prior problem, these approaches still exhibit the following fundamental limitations: 1) they indeed leverage some visual-augmented strategies, such as extra visual annotations and visual explanations derived from the gradient information, to effectively reduce the language reliance. However, these methods cannot sufficiently ascertain how these visual cues explicitly affect the learned answer representation, which demands further exploration.
2) Beyond that, existing methods generally emphasize the inter-class discrimination towards the answer representation, which overlooks more fine-grained intrinstic patterns, such as inter-instance discrimination or/and intra-instance invariance, and may hence yield unsatisfactory performance.

To tackle the aforementioned two issues, in this paper, we present a visual perturbation-aware collaborative learning framework, to explicitly and reasonably stipulate the visual effects towards the answer representation based on the intra-instance invariance and the inter-instance discrimination strategies, thereby effectively overcoming the language prior problem. The rationale of visual perturbation learning is illustrated in Fig.~\ref{fig:vpl}. Particularly, the prescribed visual effects are conducted on two new curated images with different perturbation extents, based on which we further investigate their intrinstic distribution patterns in the latent representation space. To achieve this, we intentionally deploy four reciprocal components. Specifically, in the visual perturbation controller, two sorts of manually-curated image features with different perturbation extents are automatically constructed on the basis of the original one. After being respectively encoded by the VQA base model, these three representations are selectively delivered to two following discriminators. When encountering the $\textit{hard-perturbation}$ image feature (false) and original ones (true), the class-aware discriminator is responsible to capture the inter-instance discrimination by distinguishing their semantic discrepancy. As an ancillary component, the relation-aware discriminator engages in exploiting the coarse-grained intra-instance invariant correlation. Their collaborative learning forces the model to learn the instance-specific fine-grained visual characteristic, which is effective in discriminating the feature space and overcoming the language prior problem.
% restrained to deduce a prominent probability degradation at true class index, so as to discover which input image region features affect the performance of answer prediction and enlarge the representation discriminability compared to semantically-similar samples.
To make the learned latent representation contain the minimal sufficient information and be immuned from the input bias, we further apply a variational information bottleneck modulator to better facilitate the learning of two discriminators. It is worth noting that our visual perturbation-aware learning scheme is model-agnostic and can be easily incorporated into existing state-of-the-art VQA models to reduce the language reliance and boost their reasoning performance.

Our main contributions can be summarized as follows:
\begin{itemize}
	\item We present a simple yet effective visual perturbation-aware calibration framework for mitigating the language reliance in VQA. To the best of our knowledge, this is the first attempt to overcome the language prior problem from the view of instance-level discriminative feature representation.
	
	\item We devise four reciprocal components, including mask perturbation controller, information bottleneck modulator, class-aware discriminator, and relation-aware discriminator. The controller automatically curates the images with different perturbation extents, based on which two discriminators are applied for intra-instance invariant and inter-instance discriminative learning. Besides, to filter more inputs bias, we deploy a bottleneck to calibrate the learning of latent representations.
	
	\item Extensive experiments demonstrate that our method can achieve dominant advances on two diagnostic datasets (VQA-CP v2 and VQA-CP v1) and robust performance on the balanced dataset (VQA v2). More remarkably, compared with the base model UpDn, our learning scheme achieves striking improvements of 20\% and 25\% on VQA-CP v2 dataset and VQA-CP v1 dataset, respectively.
\end{itemize}

\section{Related Work}
In this section, we briefly review the following two research directions closely related to this work: Visual Question Answering and Language Prior Alleviation in VQA.

\subsection{Visual Question Answering}
Visual Question Answering (VQA) targets at accurately answering natural language questions regarding a visual scene, which requires thorough understanding over the two heterogeneous modalities. Existing VQA studies can be roughly divided into five categories: 1) \textit{Embedding-based methods} generally embed the visual and question features into a common space by various multimodal mathematical techniques~\cite{DBLP:conf/emnlp/FukuiPYRDR16, DBLP:conf/aaai/Ben-younesCTC19}. Later on, 2) \textit{Attention-based methods}~\cite{DBLP:conf/cvpr/00010BT0GZ18, DBLP:conf/nips/KimJZ18, DBLP:conf/cvpr/GaoJYLHWL19} are proposed to assign distinctive weights on the question words or image regions, making the answering process to be more interpretable and efficient. Concurrent with aforementioned studies, 3) \textit{Relation-based methods}~\cite{DBLP:conf/nips/Norcliffe-Brown18, DBLP:conf/iccv/LiGCL19, DBLP:conf/nips/WuLWD18} harness the visual or semantic intra-modal relations to further enrich the modality context representation. 4) \textit{Modular-based methods}~\cite{DBLP:conf/iccv/HuARDS17, DBLP:conf/cvpr/ShiZL19} are also investigated to leverage the syntax structure of question to assemble a collection of function-specific modules for explicit compositional reasoning. Lastly, it is obvious that these aforementioned models are incapable of accurately answering the question that requires external information beyond the given
image. 5) \textit{Knowledge-based methods}~\cite{DBLP:conf/nips/NarasimhanLS18, DBLP:conf/ijcai/ZhuYWS0W20, DBLP:journals/pami/WangWSDH18, DBLP:conf/cvpr/MarinoRFM19} thus are advocated to leverage information beyond the given image-question pair, such as retrieving basic factual knowledge or external commonsense knowledge from DBpedia~\cite{2007DBpedia} or ConceptNet~\cite{sharma-etal-2018-conceptual} to boost their knowledge reasoning capability. 

While these approaches have achieved promising results in VQA, nonetheless, they heavily rely on the statistical regularities of concurrence between the question type and the answer during training. They thus may achieve unexpectedly impressive results on the test set sharing the same distribution with the training set, but suffer from unsatisfactory performance on out-of-domain test sets.

\subsection{Language Prior Alleviation in VQA}
To overcome the language prior problem in VQA, numerous efforts~\cite{DBLP:conf/cvpr/AgrawalBPK18, DBLP:conf/cvpr/NiuTZL0W21, DBLP:conf/emnlp/LiangJHZ20} have been dedicated to the following three respects: 1) \textit{Methods in the Architecture Side.} Most literatures of this line~\cite{DBLP:conf/eccv/KVM20, DBLP:conf/sigir/Liang0Z21, DBLP:conf/nips/RamakrishnanAL18, DBLP:conf/nips/CadeneDBCP19} make efforts to directly design explicit debiasing mechanisms. As a canonical diagnostic model, GVQA ~\cite{DBLP:conf/cvpr/AgrawalBPK18, DBLP:conf/cvpr/ShiZL19} explicitly disentangles the recognition of visual concepts from the identification of plausible answer space, which offers better generalizability across different distributions of answers. Following the same philosophy, DLR~\cite{DBLP:conf/cvpr/ShiZL19} decouples the language-based concept discovery and vision-based concept verification to reduce the question dependency. To make the question representation obtain sufficient visual-grounding, VGQE~\cite{DBLP:conf/eccv/KVM20} utilizes both visual and language modalities equally while encoding the question. Different from above one-branch methods, a line of work~\cite{DBLP:conf/sigir/Liang0Z21, DBLP:conf/nips/CadeneDBCP19, DBLP:conf/nips/RamakrishnanAL18} intentionally introduces a separated QA branch to capture the language prior on the training set in an adversarial manner~\cite{DBLP:conf/nips/RamakrishnanAL18} or a multi-task learning manner. In~\cite{DBLP:conf/nips/CadeneDBCP19}, the output of the base VQA branch is re-calibrated by utilizing element-wise multiplication operation with the mask that derived from the question-only branch for the language prior alleviation. Similarly, LPF~\cite{DBLP:conf/sigir/Liang0Z21} further casts a modulating factor from the question-only branch as a self-adaptive weight to each training sample, whereby reshaping the total VQA loss to a more balanced form. Ramakrishnan \textit{et al.}~\cite{DBLP:conf/nips/RamakrishnanAL18} set the base VQA branch and the question-only branch adversary against each other to reduce the language dependency. With the prevailing advent of the casual inference, Niu \textit{et al.}~\cite{DBLP:conf/cvpr/NiuTZL0W21} proposed a novel cause-effect look at the language bias, where a three-branch architecture is designed inspired by the view of causal graph~\cite{Morgan2007Counterfactuals}. 2) \textit{Methods in the Data Augmentation side.} One straightforward solution is to carefully construct more balanced datasets to reduce the priors. For example,  HINT~\cite{DBLP:conf/iccv/SelvarajuLSJGHB19} rigorously improves the image dependency via aligning visual importance map and extra provided human-attention maps. Analogously, SCR~\cite{DBLP:conf/nips/WuM19} introduces a training objective that ensures the visual explanations of correct answers match the most influential image regions compared with other candidate answers. To reduce the dependency on such cumbersome human annotations, several efforts have been proposed. For example, CSS~\cite{DBLP:conf/cvpr/0016YXZPZ20} encourages the model to accurately answer questions by adding complementary samples which are generated by masking objects in the image or some keywords in the question. Zhu \textit{et al.}~\cite{DBLP:conf/ijcai/ZhuMLZWZ20} incorporated an auxiliary supervised task into the base VQA model by feeding additional irrelevant pairs and lowering the prediction score of the correct answer when no supporting visual evidence is provided. More flexibly, AcSeek~\cite{DBLP:conf/cvpr/TeneyH19} first dynamically retrieves a large set of relevant questions/answers or images/captions to the given instance from an external resource, and adopts cross-domain adaption scheme to further improve the generalizability of model via gradient-based meta learning. And 3) \textit{Methods in the Loss Design side.} Apart from above two sides, some efforts have focused on designing the light-weight regularization losses for prior alleviation. For example, Liang \textit{et al.}~\cite{DBLP:conf/emnlp/LiangJHZ20} introduced a novel self-supervised contrastive learning scheme to learn the relationship between the complementary samples generated by the CSS~\cite{DBLP:conf/cvpr/0016YXZPZ20} component, and assist the model to reduce the language reliance. From the view of answer feature space learning, Guo \textit{et al.}~\cite{DBLP:conf/ijcai/GuoNCJZB21} addressed the problem by introducing an adapted margin cosine loss for different answers in the angular space under the corresponding question type, which effectively separates the answer embeddings and refrains the linguistic preference among answer types. To formulate a more general-purpose scheme, Guo \textit{et al.}~\cite{DBLP:journals/tip/GuoNCTZ22} provided an interpretation for the language prior problem from a class-imbalance view, and appropriately adopt a loss re-scaling approach to assign distinctive weights to each answer according to the training data statistics for estimating the final loss.

In our work, we mainly focused on how to effectively leverage the visual clues to reduce the language reliance. A large amount of approaches~\cite{DBLP:conf/nips/RamakrishnanAL18, DBLP:conf/emnlp/LiangJHZ20, DBLP:conf/iccv/SelvarajuLSJGHB19, DBLP:conf/nips/WuM19, DBLP:conf/cvpr/0016YXZPZ20, DBLP:conf/aaai/PatroAN20, DBLP:conf/iccv/SelvarajuCDVPB17, DBLP:conf/ijcai/ZhuMLZWZ20} have been devoted to it, yet they are deficient to investigate how the fine-grained visual perturbation explicitly affect the answer representation, which demands for further exploration. Beyond that, they generally adopt the class-level discrimination regularization towards the learned answer representation, which may fail to capture the more fine-grained instance-level characteristic for better language problem alleviation.
\vspace{-1em}
\section{Our Proposed Method}
\subsection{Problem Formulation}
The VQA task targets at automatically answering textual questions according to the given image. Formally, given a batch of data samples consisting of $B$ triplets of images $\mathcal{V}_{i}$, question $\mathcal{Q}_{i}$ and ground-truth answer set $\mathcal{A}_{i}$, denoted as $\mathcal{B} = \{\mathcal{V}_{i}, \mathcal{Q}_{i}, \mathcal{A}_{i} \}^{B}_{i=1}$, the VQA model aims to learn a mapping function $\mathcal{H}_{vqa}: \mathcal{V} \times \mathcal{Q} \rightarrow \mathcal{A}$ to produce an accurate answer. It typically involves three parts: Visual and Textual Encoder, VQA Base Model, and Answer Classifier. As the visual and textual encoder, each image $\mathcal{V}_{i}$ is encoded as the visual embedding matrix $\mathbf{V}_{i} \in \mathbb{R}^{ d_{v} \times N}$ by a pre-trained Faster-RCNN model~\cite{DBLP:conf/nips/RenHGS15} $\mathcal{U}_{v}$, where $N$ is the number of detected objects and each column represents one image region feature. For each input question $\mathcal{Q}_{i}$, the question encoder $\mathcal{U}_{q}$ is adopted to generate the question representation $\mathbf{q}_{i} \in \mathbb{R}^{d_{q}}$,
\begin{equation}
    \label{eq:1}
    \left\{ \begin{array}{ll} 
       \mathbf{V}_{i} = \mathcal{U}_{v}(\mathcal{V}_{i}), \\ 
       \mathbf{q}_{i} = \mathcal{U}_{q}(\mathcal{Q}_{i}).
\end{array} \right.
\end{equation}
After feature extraction, the VQA base model embeds the two features as a common answer representation by multi-modal fusion techniques,
\begin{equation}
	\begin{aligned}
    \label{eq:2}
    \mathbf{m}_{i} = \mathcal{H}_{m}(\mathbf{V}_{i}, \mathbf{q}_{i}; \bm{\Theta}_{m}),
    \end{aligned}
\end{equation}
where $\mathbf{m}_{i}$ denotes the answer representation of the $i$-th instance, and $\bm{\Theta}_{m}$ denotes the parameters of model $\mathcal{H}_{m}$. Following the prevailing formulation, we consider the VQA task as an orthodox multi-label classification problem. As such, the objective function is depicted as, 
\begin{equation}
	\begin{aligned}
    \label{eq:3}
    \hat{a} = \arg \max_{a_{k} \in \bm{\Omega}} p(a_{k}|\mathbf{m}_{i};\bm{\Theta}_{c}),
    \end{aligned}
\end{equation}
where $\bm{\Theta}_{c}$ represents the parameters of the answer classifier, and $a_{k}$ denotes the $k$-th answer in the answer set $\bm\Omega$. Note that each instance probably has multiple correct answers, hence the whole model is typically optimized by minimizing the following loss,
\begin{equation}
	\begin{aligned}
    \label{eq:4}
    \mathcal{L}_{vqa} = -\frac{1}{B}\sum_{i=1}^{B}\sum_{k=1}^{\left| \bm{\Omega} \right|}y_{i,k}~ \mathrm{log}(p(a_{k}|\mathbf{m}_{i};\bm{\Theta}_{c})),
    \end{aligned}
\end{equation}
where $y_{i,k}$ is the soft score of the $k$-th answer for the $i$-th instance, which is calculated by $y_{i,k} = \frac{\# vote}{\left |\mathcal{A}_{i} \right|}$, $\left |\mathcal{A}_{i} \right|$ is the number of the provided ground-truth answers for that question, and $\# vote$ denotes the number of human answer $a_{k}$ to that question.

\begin{figure*}
\hspace*{-2em}
 \includegraphics[width=190mm]{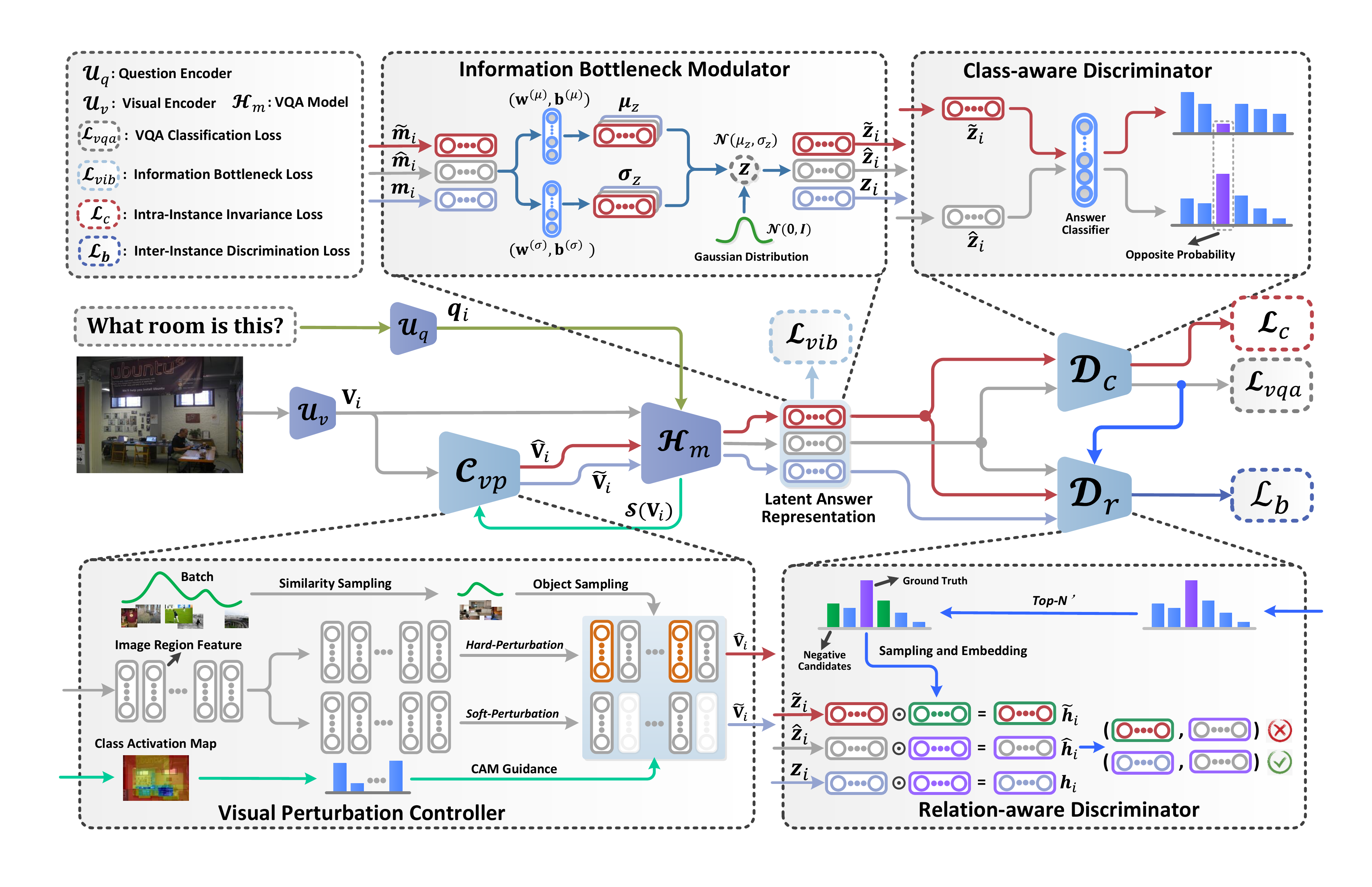}
 \vspace{-7ex}
 \caption{Schematic illustration of our proposed model. The image and question are first encoded by $\mathcal{U}_{v}$ and $\mathcal{U}_{q}$, respectively. After that, the extracted visual feature $\mathbf{V}_{i}$ is delivered to the visual-perturbation controller $\mathcal{C}_{vp}$ to generate two new perturbed visual features $\mathbf{\tilde{V}_{i}}$ and $\mathbf{\hat{V}}_{i}$. After encoding by the base VQA model $\mathcal{H}_{m}$, three sorts of latent answer representations are further calibrated to filter the redundant information by the information bottleneck modulator. Finally, two discriminators, namely, class-aware discriminator and relation-aware discriminator, are intentionally deployed for jointly optimizing the final answer representation. Beyond that, four regularization losses are adopted to optimize the overall network.}
 \vspace{-2ex}
 \label{fig:framework}
\end{figure*}

\subsection{Visual Perturbation-aware Collaborative Learning}\label{subsecvrep}
In order to explicitly investigate the implication of visual clues on the representation learning and language prior alleviation, in this work, we present a visual perturbation-aware collaboration scheme, where the intra-instance invariant and inter-instance discriminative learning are jointly considered. As illustrated in Fig.~\ref{fig:framework}, our learning scheme consists of four components: 1) Visual Perturbation Controller, 2) Information Bottleneck Modulator, 3) Class-aware Discriminator, and 4) Relation-aware Discriminator. The Visual Perturbation Controller first constructs two sorts of perturbed image features based on the original image feature, where these features along with the original one are encoded as the latent question-guided representations. Subsequently, they are delivered to the Class-aware Discriminator and Relation-aware Discriminator for inter-instance discriminative and intra-instance invariant representation learning, respectively. Moreover, the information bottleneck modulator is further employed to impose a more robust representation prior. In what follows, we will elaborate each component and how they systematically work as a whole to ameliorate the learning of the VQA base model.
% \begin{itemize}
% 	\item \textit{Perturbation-based Visual Calibration} comprises of two components: 1) we intentionally substitute each image regions with patches stemmed from the semantically similar samples, and then force the base VQA model to excavate the curated mendacious properties in an adversarial manner, thereby outputting a sufficiently discriminative answer feature for prediction. To futher improve the expressiveness of the answer representation, 2) we restrict the coarse-grained relationship between two types of  manually-curated images with the guidance of answer semantic in a contrastive manner.
% 	\item \textit{VIB-Based Semantic Refinement} engages in preserving the minimal sufficient information regarding the input image-question pairs, which is responsible to refine the answer representation for prediction.
% \end{itemize}

\textbf{Visual Perturbation Controller} ($\mathcal{C}_{vp}$). As emphasized previously, the implication of visual clues on the learned representation should be explicitly and systematically investigated. In this work, we delve into how the visual perturbation affects the expressiveness of the answer representation. As a preparation, we introduce two sorts of perturbation strategies, namely, \textit{soft-perturbation} and \textit{hard-perturbation}. The former maintains the crucial region features of the original image and only discards some marginal clues, while the latter is largely deprived of crucial information compared to the original one. 
% \footnote{We indeed can tailor more diverse perturbation to adapt to different task.}
Next, we introduce how to construct these two sorts of perturbation image features. 

Specifically, we first apply the modified Grad-CAM~\cite{DBLP:conf/iccv/SelvarajuCDVPB17} to obtain the initial contribution score of the $n$-th region feature in $\mathbf{V}_{i}$ to the answers in $\bm\Omega$ by the following function,
\begin{equation}
    \label{eq:5}
     \mathcal{S}(\mathbf{V}_{i}) = \sum_{k=1}^{\left| \bm{\Omega} \right|}((\nabla_{\mathbf{V}_{i}} p(a_{k}|\mathbf{m}_{i};\bm{\Theta}_{c}))^\top\mathbf{1}),
\end{equation}
where $\nabla_{\mathbf{V}_{i}}$ represents the gradient operation with respect to the visual embedding matrix $\mathbf{V}_{i}$, $\mathbf{1}$ denotes the vector with all elements one, and $\mathcal{S}(\mathbf{V}_{i})$ denotes the visual contribution score vector for the $i$-th image, each element of which represents the contribution score of each region. We dynamically collect several region features with top score in $\mathcal{S}(\mathbf{V}_{i})$ and tentatively treat them as the salient ones regarding the $i$-th image, denoted as $\mathbf{O}_{i}^{*}$. To construct the hard-perturbation image, we mask these salient object features with the substitudes (i.e., new region features) from other instances. Thereafter, we intentionally develop a substitude feature distribution $\mathbf{V}_{j} \sim g(\mathbf{V}|\mathcal{B}, R_{i,j})$, wherein the instance specified by each feature $\mathbf{V}_{j}$ is required to be semantically similar to the $i$-th instance and be part of the same training batch $\mathcal{B}$ as the $i$-th instance. $R_{i,j}$ is described as the correlation between the $i$-th instance and the $j$-th instance, which is computed by the cosine function as follows,
\begin{equation}
    \label{eq:6}
    R_{i,j} = \mathrm{cos}(\mathcal{P}(\mathbf{V}_{i}) \odot \mathbf{q}_{i}, \mathcal{P}(\mathbf{V}_{j}) \odot \mathbf{q}_{j}),
\end{equation}
where $\mathcal{P}(\cdot)$ and $\odot$ denote the mean pooling operation and hadamart product, respectively. $\mathcal{P}(\mathbf{V}_{i}) \odot \mathbf{q}_{i}$ denotes the joint vision-linguistic feature of the $i$-th instance. After constructing the distribution $\mathbf{V}_{j} \sim g(\mathbf{V}|\mathcal{B}, R_{i,j})$, we further randomly sample $K$ image region features from a sampled image feature $\mathbf{V}_{j}$, described as $\mathbf{O}_{i}^{s} \sim t(\mathbf{V}_{j})$, and substitute the salient object feature $\mathbf{O}_{i}^{*}$ of $i$-th instance with $\mathbf{O}_{i}^{s}$ as,
\begin{equation}
    \label{eq:7}
    \mathbf{\hat{V}}_{i} = \mathcal{M}(\mathbf{O}_{i}^{*}, \mathbf{O}_{i}^{s}, K),
\end{equation}
where $\mathcal{M}(\cdot)$ denotes the mask operation, $K$ ($K \leq \left| \mathbf{O}_{i}^{*} \right|$) controls the number of region features for mask and is dynamically adjusted with training epochs. By means of Eq.~(\ref{eq:7}), the $\textit{hard-perturbation}$ image feature $\mathbf{\hat{V}}_{i}$ is constructed. For the $\textit{soft-perturbation}$ image feature, we mask some trivial object features in $\mathbf{V}_{i}$ with zero mask, which is only slightly disturbed and distinctly differs from the hard-perturbation one,
\begin{equation}
    \label{eq:8}
    \mathbf{\tilde{V}}_{i} = \mathcal{M}(\mathbf{V}_{i} \backslash \mathbf{O}_{i}^{*}, \mathbf{0}, p - K),
\end{equation}
where $\mathbf{V}_{i} \backslash \mathbf{O}_{i}^{*}$ represents the remnant region features with low contribution score in $\mathcal{S} (\mathbf{V}_{i})$, $\mathbf{0}$ denotes the mask with zero elements, and $p$ ($p \leq \left| \mathbf{V}_{i} \backslash \mathbf{O}_{i}^{*} \right|$ + $K$) is the number of attended regions.

Based on these two newly constructed image features, we tentatively develop two intuitive principles as follows, 
\begin{itemize}
	\item (\textit{Principle I}) The soft perturbation on the trivial region features of each image would result in a similar representation compared to that of the original image in the relation-aware space.
	\item (\textit{Principle II}) The hard perturbation on the salient region features of each image would result in a semantically-countered answer representation compared to that of the original image in the class-aware space.
\end{itemize}

Following these principles, we subsequently devise two discriminators to make the best of visual perturbation features.

\textbf{Relation-aware Discriminator} ($\mathcal{D}_{r}$). According to \textit{Principle I}, we propose the relation-aware discriminator to investigate the impact of soft perturbation on the learned answer representation, and develop the intra-instance invariant loss under the semantic guidance of the answer embedding. 

Specifically, we first select the top-$N^{'}$ negative candidate answers $\left \{ a_{k} \right \}_{k=1}^{N^{'}}$ with the highest probability score from the initial model prediction distribution, and these candidate answers along with the ground-truth answers are respectively embedded as the negative embeddings $\left \{ \mathbf{a}_{k} \right \}_{k=1}^{N^{'}}$ and the ground-truth embedding $\mathbf{A}_{i}^{gt}$ using the question encoder $\mathcal{U}_{q}$, where each row of $\mathbf{A}_{i}^{gt}$ represents the embedding of one of the ground-truth answers for the $i$-th instance. 

Next, we generate three sorts of new latent representations by taking both the question-guided representation and the answer semantics into consideration,
\begin{equation}
    \label{eq:9}
    \left\{ \begin{array}{ll} 
       \mathbf{h}_{i} = \mathbf{m}_{i} \odot f_{a}(\mathcal{P}(\mathbf{A}_{i}^{gt}); \bm{\Theta}_{a}), \\ 
       \mathbf{\tilde{h}}_{i} = \mathbf{\tilde{m}}_{i} \odot f_{a}(\mathcal{P}(\mathbf{A}_{i}^{gt}));\bm{\Theta}_{a}), \\
       \mathbf{\hat{h}}_{i} = \mathbf{\hat{m}}_{i} \odot f_{a}(\mathbf{a}_{c};\bm{\Theta}_{a}),
\end{array} \right.
\end{equation}
where $\mathbf{a}_{c}$ is sampled from the negative candidate answer distribution, i.e., $\mathbf{a}_{c} \sim p_{a}(\mathbf{a}|\left \{ \mathbf{a}_{k} \right \}_{k=1}^{N^{'}})$, $\mathbf{\hat{m}}_{i}$ and $\mathbf{\tilde{m}}_{i}$ denote the encoded representation of hard-perturbation features $\mathbf{\hat{V}}_{i}$ and soft-perturbation features $\mathbf{\tilde{V}}_{i}$, respectively. $f_{a}(\cdot;\bm{\Theta}_{a})$ denotes the mapping function from the answer space to the vision-linguistic space, and $\mathcal{P}$ denotes the mean pooling operation. Intuitively, the joint representation $\mathbf{\tilde{h}}_{i}$ should be rigorously consistent with the original one $\mathbf{h}_{i}$, while evidently distinguishes the \textit{hard-perturbation} one $\mathbf{\hat{m}}_{i}$. This is due to the fact that \textit{soft-perturbation} representation $\mathbf{\tilde{h}}_{i}$ still preserves the crucial region information of $\mathbf{h}_{i}$, and the \textit{hard-perturbation} one $\mathbf{\hat{{h}}_{i}}$ degenerates as a result of the loss of important visual clues as well as the affect of the negative answer semantic.

In light of this, we hereby formulate the intra-instance invariant loss as follows,
% We take the cosine similarity of the representations in the joint embedding space as our scoring function because it implicitly normalizes the embeddings.
\begin{equation}
    \label{eq:10}
	\begin{aligned}
     \mathcal{L}_{b} = \frac{1}{B}\sum_{i=1}^{B} \mathrm{log}(2- \frac{e^{\mathrm{cos}(\mathbf{h}_{i}, \mathbf{\tilde{h}}_{i})}}{e^{\mathrm{cos}(\mathbf{h}_{i}, \mathbf{\tilde{h}}_{i})}+e^{\mathrm{cos}(\mathbf{\tilde{h}}_{i}, \mathbf{\hat{h}}_{i})}}).
    \end{aligned}
\end{equation}
While facing the soft-perturbation of the original region features, maintaining the intra-instance invariance enables the model to immune to the alteration of trivial visual information, which could better capture the crucial instance-level visual characteristic and alleviate the language prior problem.
% It is noteworthy that minimizing $-\mathrm{log}(\mathbf{x})$ is equivalent to minimizing $\mathrm{log}(2-\mathbf{x})$ mathematically, where $\mathbf{x}$ denotes any input vector. Experimentally, minimizing $\mathrm{log}(2-\mathbf{x})$ is more stable than minimizing $-\mathrm{log}(\mathbf{x})$ during training, thus we adopt the latter form as the final regularization.

\textbf{Class-aware Discriminator} ($\mathcal{D}_{c}$). As a semantic-level discriminator, the class-aware discriminator is required to plainly distinguish the semantic discrepancy between the $\textit{hard-perturbation}$ representation $\mathbf{\hat{m}}_{i}$ and the original one $\mathbf{m}_{i}$ following \textit{Principle II}. Based on this consideration, the inter-instance discrimination regularization loss is formulated to broader the instance-level representation space and reduce the affect of the language reliance,
\begin{equation}
	\begin{aligned}
    \label{eq:11}
    \mathcal{L}_{c} = \frac{1}{B}\sum_{i=1}^{B}\sum_{k=1}^{\left| \bm{\Omega} \right|}p(a_{k}|\mathbf{m}_{i};\bm{\Theta}_{c})~ \mathrm{log}(p(a_{k}|\mathbf{\hat{m}}_{i};\bm{\Theta}_{c}))),
    \end{aligned}
\end{equation}
where $\bm{\Omega}$ and $\bm{\Theta}_{c}$ denote the answer set and the parameter of the classifier, respectively. $p(a_{k}|\mathbf{m}_{i};\bm{\Theta}_{c})$ and $p(a_{k}|\mathbf{\hat{m}}_{i};\bm{\Theta}_{c})$ are the predictions on $a_{k}$ given the original representation $\mathbf{m}_{i}$ and the hard-perturbation representation $\mathbf{\hat{m}}_{i}$, respectively. By minimizing the $\mathcal{L}_{c}$ loss, these two prediction distributions are constrained to be opposite. Upon deeper analysis, when perturbed by other instances, the opposite constraint in the semantic space (i.e., class prediction distribution) could help better distinguish which image regions are more faithful to model decisions and capture the instance-specific fine-grained visual clues, therefore inhibit the dominant impact of language prior.

\textbf{Information Bottleneck Modulator}. The two tailored discriminators as well as the supervision of the intra-instance and inter-instance regularization guarantee the explicit visual perturbation effects and the instance-level characteristic learning. However, the latent answer representation encoded by the VQA base model may still involve some redundant information (i.e., input bias), which hinders the feature discriminability and weaken the generalizability of the VQA encoder. 

As the variational information bottleneck~\cite{DBLP:conf/iclr/AlemiFD017} could enforce the representation to preserve the minimal sufficient information flow, we intentionally deploy an information bottleneck modulator between the VQA model and two discriminators to amputate the redundant noise, and the regularization loss is depicted as the Kullback-Leibler (KL) divergence between two distributions $p_{z}(\mathbf{z}_{i}|\mathbf{m}_{i};\bm{\Theta}_{vib})$ and $r(\mathbf{z}_{i})$,
\begin{equation}
	\begin{aligned}
    \label{eq:12}
		\mathcal{L}_{vib} = \frac{1}{B}\sum_{i=1}^{B}D_{KL}(p_{z}(\mathbf{z}_{i}|\mathbf{m}_{i};\bm{\Theta}_{vib})~||~r(\mathbf{z}_{i})),
    \end{aligned}
\end{equation}
where $\mathbf{z}_{i}$ is the latent encoding variable of the $i$-th instance, $p_{z}(\mathbf{z}_{i}|\mathbf{m}_{i})$ is an encoded distribution conditioned on the answer representation $\mathbf{m}_{i}$, and $r(\mathbf{z}_{i})$ is a standard normal distribution with the same dimension as $\mathbf{z}_{i}$. To further achieve the encoding process of $p_{z}(\mathbf{z}_{i}|\mathbf{m}_{i};\bm{\Theta}_{vib})$, we sample from the posterior $\mathbf{z}_{i}$ via the reparameterized trick~\cite{DBLP:conf/iclr/AlemiFD017},
\begin{gather}
	\label{eq:13}
	\begin{cases}
		\bm{\mu}_{i} = \mathbf{W}^{(\mu)}\mathbf{m}_{i}+\mathbf{b}^{(\mu)},\\
		\bm{\sigma}_{i} = \mathbf{W}^{(\sigma)}\mathbf{m}_{i}+\mathbf{b}^{(\sigma)},\\
		\mathbf{z}_{i} = \bm{\mu}_{i} + \bm{\epsilon}  \odot \bm{\sigma}_{i} \sim  p_{z}(\mathbf{z}_{i}|\mathbf{m}_{i};\bm{\Theta}_{vib}),\\
	\end{cases}
\end{gather}
 where $\mathbf{W}^{(\sigma)}$, $\mathbf{W}^{(\mu)}$, $\mathbf{b}^{(\sigma)}$, and $\mathbf{b}^{(\mu)}$ are the learnable weights, the mean vector $\bm{\mu}_{i}$ and the variance vector $\bm{\sigma}_{i}$ of each instance are encoded with two fully connected layers, and $\bm{\epsilon} \sim \mathcal{N}(\mathbf{0}, \mathbf{I})$ is a noisy variable sampled from an isotropic multivariate Gaussian distribution. By means of the regularization of Eq.~(\ref{eq:12}), the obtained representation $\mathbf{z}_{i}$ is more robust compared to the original one $\mathbf{m}_{i}$, which could further benefit the joint learning of the intra-instance invariance and the inter-instance discrimination for language prior alleviation. 
 \vspace{-1em}
\subsection{Collaborative Learning and Model Optimization}
Finally, we combine these components in a collaborative framework, which could systematically work as a whole and reciprocate each other for robust representation learning and language prior reduction. As such, the weighted summation of the three regularization losses and the base VQA classification loss make up the overall loss,
\begin{equation}
    \label{eq:14}
     \mathcal{L}_{total} = \mathcal{L}_{vqa}  + \lambda_{vib}\mathcal{L}_{vib} + \lambda_{b}\mathcal{L}_{b} + \lambda_{c}\mathcal{L}_{c},
\end{equation}
where $\lambda_{vib}$, $\lambda_{b}$, and $\lambda_{c}$ are the hyper-parameters to control the weights of the three losses. Obviously, $\mathcal{L}_{total}$ can be dismissed as a generalized VQA loss, as it degenerates to $\mathcal{L}_{vqa}$ when $\lambda_{vib}$ = $\lambda_{b}$ = $\lambda_{c}$ = 0. Upon further analysis, $\mathcal{L}_{vib}$ actually acts as a regularizer on the latent encoded space, preventing the VQA model from memorizing the trival information of inputs. Based on this, $\mathcal{L}_{b}$ and $\mathcal{L}_{c}$ further optimize the base model by harnessing the combination of intra-instance variant and inter-instance discriminative learning. Notably, our proposed method supports model-agnostic setting, which means that it can be flexibly plugged into most existing methods to overcome the language prior problem. Besides, the visual perturbation-aware scheme introduces little additional parameters, and the increased computational cost is thus negligible.

For better illustrating our framework, we hereby summarize the training process and formulate the overall algorithm pipeline. Specifically, we first merge the training parameters based on whether they are subordinate to the same loss branch. Thereafter, they are reorganized as the following form, 
\begin{equation}
    \label{eq:15}
    \left\{ \begin{array}{ll} 
       \mathcal{L}_{vqa}:\bm{\Theta}_{\mathcal{L}_{vqa}} = \left \{ \bm{\Theta}_{m}, \bm{\Theta}_{vib}, \bm{\Theta}_{c}\right \}, \\ 
       \mathcal{L}_{vib}:\bm{\Theta}_{\mathcal{L}_{vib}} = \left \{ \bm{\Theta}_{m}, \bm{\Theta}_{vib} \right\}, \\
       \mathcal{L}_{c}:\bm{\Theta}_{\mathcal{L}_{c}} = \left \{ \bm{\Theta}_{m}, \bm{\Theta}_{vib}, \bm{\Theta}_{c} \right\}, \\
       \mathcal{L}_{b}:\bm{\Theta}_{\mathcal{L}_{b}} = \left \{ \bm{\Theta}_{m}, \bm{\Theta}_{vib}, \bm{\Theta}_{a} \right\}.
\end{array} \right.
\end{equation}
In the training procedure, we first pretrain the model with the orthodox VQA loss $\mathcal{L}_{vqa}$ and information bottleneck loss $\mathcal{L}_{vib}$ for several epochs, and the derivative of $\mathcal{L}_{(vqa,vib)} = \mathcal{L}_{vqa} + \lambda_{vib}\mathcal{L}_{vib}$ with respect to $\bm{\Theta}_{vqa}$ and $\bm{\Theta}_{vib}$ can be calculated,
\begin{equation}
    \label{eq:16}
    \left\{ \begin{array}{ll} 
       \bm{\Theta}_{vqa} = \bm{\Theta}_{vqa} - \eta \frac{\partial \mathcal{L}_{(vqa, vib)}}{\partial \bm{\Theta}_{vqa}}, \\ 
       \frac{\partial \mathcal{L}_{(vqa, vib)}}{\partial \bm{\Theta}_{vqa}} = \frac{\partial \mathcal{L}_{vqa}}{\partial \bm{\Theta}_{vqa}} + \lambda_{vib} \frac{\partial \mathcal{L}_{vib}}{\partial \bm{\Theta}_{vib}}, \\
       \bm{\Theta}_{vib} = \bm{\Theta}_{vib} - \eta \frac{\partial \mathcal{L}_{(vqa, vib)}}{\partial \bm{\Theta}_{vib}}, \\ 
       \frac{\partial \mathcal{L}_{(vqa, vib)}}{\partial \bm{\Theta}_{vib}} = \frac{\partial \mathcal{L}_{vqa}}{\partial \bm{\Theta}_{vib}} + \lambda_{vib} \frac{\partial \mathcal{L}_{vib}}{\partial \bm{\Theta}_{vib}},
\end{array} \right.
\end{equation}
where $\eta$ denotes the step size. After pretraining, we continually finetune the model by adding other two losses: the intra-instance invariance loss $\mathcal{L}_{b}$ and the inter-instance discrimination loss $\mathcal{L}_{c}$. The detail of the derivative form follows Eq.~(\ref{eq:16}), and the overall algorithm pipeline is summarized in Algorithm~\ref{algorithm}.
\begin{algorithm}[t]
 \caption{The pipeline of optimizing our visual perturbation-aware collaborative framework.}
 \label{algorithm}
 \begin{algorithmic}[1]
  \REQUIRE Image feature $\mathbf{V}_{i}$; Question feature $\mathbf{q}_{i}$; Ground-truth answer embeddings $\mathbf{A}_{i}^{gt}$; $\lambda_{vib}$, $\lambda_{c}$, $\lambda_{b}$; $T_{0}$, $T_{1}$, $T_{2}$. \\
  \ENSURE The parameters of $\bm{\Theta}_{m}$,  $\bm{\Theta}_{vib}$, $\bm{\Theta}_{c}$, and $\bm{\Theta}_{a}$.
  \STATE Randomly initialize all the parameters.
  \FOR{$t$ in $T_{0}$ epochs}
  \STATE Obtain $\mathbf{m}_{i}$ by Eq.~(\ref{eq:2}).
  \STATE Obtain $\mathbf{z}_{i}$ by Eq.~(\ref{eq:13}).
  \STATE Pretrain by optimizing the following loss,
  \STATE \ \ \ \ \ \ \ \ \ \ \ \ \ \ \ \ $\mathcal{L}_{(vqa, vib)}$ = $\mathcal{L}_{vqa}$ + $\lambda_{vib} \mathcal{L}_{vib}$.
  \ENDFOR
  \FOR{$t$ in $T_{1}-T_{0}$ epochs}
  \STATE Obtain $\mathbf{\hat{V}}_{i}$ and $\mathbf{\tilde{V}}_{i}$ by Eq.~(\ref{eq:7}) and Eq.~(\ref{eq:8}), respectively.
  \STATE Obtain $\mathbf{m}_{i}$, $\mathbf{\hat{m}}_{i}$, and $\mathbf{\tilde{m}_{i}}$ by Eq.~(\ref{eq:2}).
  \STATE Obtain $\mathbf{z}_{i}$, $\mathbf{\hat{z}}_{i}$, and $\mathbf{\tilde{z}_{i}}$ by Eq.~(\ref{eq:13}).
  \STATE Finetune by optimizing the following loss,
  \STATE \ \ \ \ \ \ \ \ \ \ $\mathcal{L}_{(vqa, vib, c)}$ = $\mathcal{L}_{vqa}$ + $\lambda_{vib} \mathcal{L}_{vib}$ + $\lambda_{c}\mathcal{L}_{c}$.
  \ENDFOR
  \FOR{$t$ in $T_{2} - T_{1}$ epochs}
  \STATE Obtain $\mathbf{\hat{V}}_{i}$ and $\mathbf{\tilde{V}}_{i}$ by Eq.~(\ref{eq:7}) and Eq.~(\ref{eq:8}), respectively.
  \STATE Obtain $\mathbf{m}_{i}$, $\mathbf{\hat{m}}_{i}$, and $\mathbf{\tilde{m}_{i}}$ by Eq.~(\ref{eq:2}).
  \STATE Obtain $\mathbf{z}_{i}$, $\mathbf{\hat{z}}_{i}$, and $\mathbf{\tilde{z}}_{i}$ by Eq.~(\ref{eq:13}).
  \STATE Obtain the triplets $\mathbf{h}_{i}$, $\mathbf{\hat{h}}_{i}$, and $\mathbf{\tilde{h}}_{i}$ by Eq.~(\ref{eq:9}).
  \STATE Finetune by optimizing the following loss,
  \STATE \ \ \ \ \ \ \ \ \ \ $\mathcal{L}_{total}$ = $\mathcal{L}_{vqa}$ + $\lambda_{vib} \mathcal{L}_{vib}$ + $\lambda_{b}\mathcal{L}_{b}$ + $\lambda_{c}\mathcal{L}_{c}$.
  \ENDFOR
  \STATE \textbf{return} $\bm{\Theta}_{m}$,  $\bm{\Theta}_{vib}$, and $\bm{\Theta}_{c}$. 
 \end{algorithmic}
\end{algorithm}
\vspace{-1em}
\subsection{Inference}
In this subsection, we elaborate the inference procedure of our framework, which slightly differs from the training procedure. Specifically, given a test image $\mathcal{V}_{i}$ and its related question $\mathcal{Q}_{i}$, our model first generates their features by $\mathcal{U}_{v}$ and $\mathcal{U}_{q}$, i.e., $\mathbf{V}_{i}$ and $\mathbf{q}_{i}$. Afterwards, these two features are encoded  via the pretrained VQA model to obtain the intermediate representation $\mathbf{m}_{i}$. Note that, since the variance vector $\mathbf{W}^{(\sigma)} \mathbf{m}_{i} + \mathbf{b}^{(\sigma)}$ tends to capture input bias during the training process, we thus only adopt the mean vector $\mathbf{W}^{(\mu)} \mathbf{m}_{i} + \mathbf{b}^{(\mu)}$ as the unbiased estimation for the answer representation, and the final answer prediction of our model can be formulated as the following form,
\begin{equation}
	\begin{aligned}
    \label{eq:17}
    \hat{a} = \arg \max_{a_{k} \in \bm{\Omega}} p(a_{k}|\mathbf{W}^{(\mu)} \mathbf{m}_{i} + \mathbf{b}^{(\mu)};\bm{\Theta}_{c}).
    \end{aligned}
\end{equation}
\section{Experiments}
\par To thoroughly demonstrate the effectiveness of our proposed model, we carried out extensive experiments to answer the following research questions (RQs):
\begin{itemize}
	\item \textbf{RQ1}: Is our proposed method able to surpass the state-of-the-art VQA models?
	\item \textbf{RQ2}: Is each component of our model effective for boosting the performance? % ablation analysis 
	\item \textbf{RQ3}: Why does the model outperform the baselines? 
	\item \textbf{RQ4}: How do the hyper-parameters and the experimental setting affect the model performance?
	\item \textbf{RQ5}: What does the model learn?
\end{itemize}
In following subsections, we first provided the basic information of the evaluated benchmark datasets. Then we introduced the standard evaluation metric, followed by the brief introduction of a series of baselines specially designed for language prior alleviation in VQA, and we ended this section with some implementation details of our method.
\subsection{Datasets}
We evaluated our proposed method on three commonly used benchmarks, which can be divided into two groups: the diagnostic datasets~\cite{DBLP:conf/nips/RamakrishnanAL18} and the balanced dataset~\cite{DBLP:conf/iccv/AntolALMBZP15}. These datasets are elaborated as follows. 

The diagnostic datasets contain: 1) VQA-CP v2~\cite{DBLP:conf/nips/RamakrishnanAL18}, with $\sim$658K visual questions of three types (\textit{Num.}, \textit{Yes/No} and \textit{Other}) based on $\sim$122K Microsoft COCO~\cite{DBLP:conf/eccv/LinMBHPRDZ14} images, which is derived from the VQA v2 dataset by reorganizing the \textit{train} and \textit{val} splits. The QA pairs in the training set and test set have significantly different distributions, therefore it is appropriate to evaluate the capability of the model to overcome the language prior problem. 2) VQA-CP v1~\cite{DBLP:conf/iccv/AntolALMBZP15}, curated from the VQA v1 dataset, involves $\sim$122K images and $\sim$370K questions, which is also developed to evaluate the generalization ability. 

As to the balanced dataset VQA v2, it is the most commonly used VQA benchmark dataset. Its images are the same as VQA-CP v2, and the distribution difference of questions between the \textit{train} set and \textit{test} set is much less. Besides, the dataset is split into \textit{train}, \textit{val}, and \textit{test} (or \textit{test-std}) splits, and 25 \% of the \textit{test-std} set is reserved as the \textit{test-dev} set.
\subsection{Baselines}
To verify the effectiveness and generalization of our proposed model in two groups of VQA datasets, we compared it against the following baselines:
\begin{itemize}
    \item Methods in the architecture side, namely, UpDn~\cite{DBLP:conf/cvpr/00010BT0GZ18}, GVQA~\cite{DBLP:conf/cvpr/AgrawalBPK18}, DLR~\cite{DBLP:conf/aaai/JingWZJW20}, VGQE~\cite{DBLP:conf/eccv/KVM20}, RUBi~\cite{DBLP:conf/nips/CadeneDBCP19}, LPF~\cite{DBLP:conf/sigir/Liang0Z21}, LMH~\cite{DBLP:conf/emnlp/ClarkYZ19}, LMH~\cite{DBLP:conf/emnlp/ClarkYZ19}, AdvReg~\cite{DBLP:conf/nips/RamakrishnanAL18}, and CF-VQA~\cite{DBLP:conf/cvpr/NiuTZL0W21};
    \item Methods in the data augmentation side, namely, HINT~\cite{DBLP:conf/iccv/SelvarajuLSJGHB19}, SCR~\cite{DBLP:conf/nips/WuM19}, SSL~\cite{DBLP:conf/ijcai/ZhuMLZWZ20}, and AcSeeK~\cite{DBLP:conf/cvpr/TeneyH19};
    \item Methods in the loss design side, namely, CSS~\cite{DBLP:conf/cvpr/0016YXZPZ20}, RaReg~\cite{DBLP:conf/nips/TeneyAKSKH20}, AdaVQA~\cite{DBLP:conf/ijcai/GuoNCJZB21}, LRS~\cite{DBLP:journals/tip/GuoNCTZ22}, and GCE-DQ~\cite{DBLP:conf/iccv/HanWSH021}.
\end{itemize}

% \begin{figure}
%   \subfigure[]{\includegraphics[width=40mm, height=31.2mm]{alpha.pdf}}
%   \subfigure[]{\includegraphics[width=40mm]{F.pdf}}
%   \vspace{-3mm}
%   \caption{(a) Performance of our LENA network with different value of scaling factor $\alpha$. (b) Number of the important regions $\Vert I(F(\mathbf{r}_{t}^{u})>0)\Vert_{0}$ at the $t$-th cell.}
%   \label{fig:alpha_and_F}
% \end{figure}

% \begin{figure}[t]
% 	\subfigure[TRT on ActivityNet Captions]{
% 		%		\label{figsrtexpansion}
% 		\includegraphics[width=0.50\textwidth]{updn.png}
% 	}
%  	\hspace{-3em}
% 	\subfigure[ART on ActivityNet Captions]{
% 		%		\label{figartexpansion}
% 		\includegraphics[width=0.50\textwidth]{luna.png}
% 	}
% % 	\vspace{-2ex}
% 	\caption{ }
% 	\label{tsne}
% \end{figure}

% \begin{figure}[t]
% 	\centering
% 	\subfloat[UpDn]{\includegraphics[width=45mm]{updn.png}
% 		\label{Fig. 1(a)}}\hspace{-1.8em}
% 	\subfloat[]{\includegraphics[width=45mm]{luna.png}
% 		\label{Fig. 2(b)}}
% 	\caption{(a) Examples of experiment 1. (b) Examples of experiment 2.}
% \end{figure}

% \floatsetup{heightadjust=all, floatrowsep=columnsep}
% \newfloatcommand{figurebox}{figure}[\nocapbeside][\dimexpr(\textwidth-\columnsep)/2\relax]
% \newfloatcommand{tablebox}{table}[\nocapbeside][\dimexpr(\textwidth-\columnsep)/2\relax]

% \begin{figure}
% 	\includegraphics[width=0.50\textwidth]{qtype_distribution.png}
% 	\caption{Performance of our LENA network with different number of LENA cells $T$ on four types of questions.}
% 	\label{fig:four_type_acc}
% \end{figure}

\subsection{Evaluation Metric}
We adopted the standard VQA accuracy metric for evaluation~\cite{DBLP:conf/iccv/AntolALMBZP15}. Given an image and a corresponding question, for a predicted answer $a$, the accuracy is computed as follows,
\begin{equation}
    \label{eq:18}
    Acc_{a} = \mathrm{min}(1, \frac{\# \mathrm{humans \ that \ provide \ \textit{a}}}{3}).
\end{equation}

It is worth noting that each question is answered by several annotators, which takes the disagreement among human answers into consideration. 
\begin{table}[t]
	\caption{Performance evaluated on the $\textit{test}$ split of VQA-CP v1.}
		\vspace{-2ex}
	\label{table:cpv1}
	\scalebox{0.98}{
		\begin{tabular}{c|c|cccc}
			\hline
			\multicolumn{1}{c|}{\multirow{2}{*}{Methods}} & \multicolumn{5}{c}{VQA-CP v1 \textit{test} Acc.(\%)}            \cr  \cline{2-6}
			\multicolumn{1}{c|}{} & Venue & All& Yes/No& Num. & Other \cr 
			\hline
			MCB~\cite{DBLP:conf/emnlp/FukuiPYRDR16} &   CVPR 2016& 34.39 & 37.96  & 11.80 & 39.90  \\ 
			SAN~\cite{DBLP:conf/cvpr/YangHGDS16} &   CVPR 2016& 26.88 & 35.34 & 11.34 & 24.70 \\ 		
			NMN~\cite{DBLP:conf/cvpr/AndreasRDK16} &   CVPR 2016& 29.64 & 38.85 & 11.23 & 27.88   \\ 
% 			HieCoAtt~\cite{DBLP:conf/nips/LuYBP16} &   CVPR' 2018&    43.87\%&  63.34\%&   53.26\%&  73.52\%  \\ 
			UpDn~\cite{DBLP:conf/cvpr/00010BT0GZ18} &   CVPR 2018& 37.96 & 42.79 & 12.41 & 42.53 \\
			Counter~\cite{DBLP:conf/iclr/ZhangHP18} &   CVPR 2018& 37.67 & 41.01 & 12.98 & 42.69  \\ 
			\hline
			GVQA~\cite{DBLP:conf/cvpr/AgrawalBPK18} &   CVPR 2018& 39.23 & 64.72 & 11.87 & 24.86 \\ 
			AdvReg~\cite{DBLP:conf/nips/RamakrishnanAL18} &   CVPR 2018&   43.43 & 74.16 & 12.44 & 25.32  \\ 
			RUBi~\cite{DBLP:conf/nips/CadeneDBCP19} &   ICCV 2019& 50.90 & 80.83 & 13.84 & 36.02 \\ 
			LMH~\cite{DBLP:conf/emnlp/ClarkYZ19} &   EMNLP 2019& 55.73 & 78.59 & 24.68 & 45.47  \\ 
			VGQE~\cite{DBLP:conf/eccv/KVM20} &   ECCV 2020& 48.75 & - & - & -  \\ 
			\hline
			SCR~\cite{DBLP:conf/nips/WuM19} &   NeurIPS 2019& 48.47 & - & - & -  \\ 
	    	SCR+HAT~\cite{DBLP:conf/nips/WuM19}  &  NeurIPS 2019& 49.17 & - & - & -  \\ 
			SSL~\cite{DBLP:conf/ijcai/ZhuMLZWZ20} &   IJCAI 2020& 60.92 & 81.97 & 41.12 & 47.53  \\
			\hline
			CSS~\cite{DBLP:conf/cvpr/0016YXZPZ20}  &  CVPR 2020& 38.36 & 43.15 & 12.79 & 43.88  \\ 
			AdaVQA~\cite{DBLP:conf/ijcai/GuoNCJZB21} &   IJCAI 2021&   61.20 & \color{blue}{\textbf{91.17}} & 41.34 & 39.38  \\ 
			\hline
			\textbf{UpDn+Ours} & - & \color{blue}{\textbf{63.03}}  & 86.19 & \color{blue}{\textbf{42.97}} & \color{blue}{\textbf{48.06}} \\ 
			\hline
		\end{tabular}
}

\end{table}
\begin{table}[t]
	\caption{Performance evaluated on the $\textit{test}$ split of VQA-CP v2.}
		\vspace{-2ex}
	\label{table:cpv2}
	\scalebox{0.98}{
		\begin{tabular}{c|c|cccc}
			\hline
			\multicolumn{1}{c|}{\multirow{2}{*}{Methods}} & \multicolumn{5}{c}{VQA-CP v2 \textit{test} Acc.(\%)}            \\ \cline{2-6}
			\multicolumn{1}{c|}{} & Venue &  All& Yes/No& Num.& Other \cr
			\hline
			SAN~\cite{DBLP:conf/cvpr/YangHGDS16} &   CVPR 2016&   24.96&  38.35&   11.14 &  21.74  \\ 
			NMN~\cite{DBLP:conf/cvpr/AndreasRDK16} &   CVPR 2016&    27.47&  38.94&  11.92&  25.72   \\ 
			MCB~\cite{DBLP:conf/emnlp/FukuiPYRDR16} &   CVPR 2016&   36.33&  41.01&   11.96&  25.72  \\ 
			HieCoAtt~\cite{DBLP:conf/nips/LuYBP16} &   CVPR 2018&    28.65&  52.25&   13.79&  20.33 \\
			UpDn~\cite{DBLP:conf/cvpr/00010BT0GZ18} &   CVPR 2018 &    39.78 &  42.98 & 12.06 & 46.50  \\ 
			Counter~\cite{DBLP:conf/iclr/ZhangHP18} &   CVPR 2018&   37.67 & 41.03 & 12.98 & 42.69  \\ 
			\hline
			GVQA~\cite{DBLP:conf/cvpr/AgrawalBPK18} &   CVPR 2018 & 31.30 & 57.99 & 13.68 & 22.14 \\ 
			AdvReg~\cite{DBLP:conf/nips/RamakrishnanAL18} &   CVPR 2018 & 41.17 & 65.49 & 15.48 & 35.48  \\
			RUBi~\cite{DBLP:conf/nips/CadeneDBCP19} &   ICCV 2019 & 47.11 & 68.65 & 20.28 & 43.18  \\ 
			LM~\cite{DBLP:conf/emnlp/ClarkYZ19} &   CVPR 2019 & 48.78 & 72.78 & 14.61 & 45.58  \\ 
			LMH~\cite{DBLP:conf/emnlp/ClarkYZ19} &   CVPR 2019 & 52.05 & 70.14 & 43.98 & 44.81  \\ 
			VGQE~\cite{DBLP:conf/eccv/KVM20} &   ECCV 2020 & 50.11 & 66.35 & 27.08 & 46.77  \\
			DLR~\cite{DBLP:conf/aaai/JingWZJW20} &  AAAI 2020& 48.87 & 70.99 & 18.72 & 45.57  \\ 
			LPF~\cite{DBLP:conf/sigir/Liang0Z21} &   SIGIR 2021 & 55.34 & 88.61 & 23.78 & 46.57  \\
 			CF-VQA~\cite{DBLP:conf/cvpr/NiuTZL0W21} & CVPR 2021 & 54.95 & \color{blue}{\textbf{90.56}} & 21.88 & 45.36 \\
 			\hline
			HINT~\cite{DBLP:conf/iccv/SelvarajuLSJGHB19} &   ICCV 2019 & 46.73 & 67.27 & 10.61 & 45.80  \\
			SCR~\cite{DBLP:conf/nips/WuM19} &   NeurIPS 2019 & 49.45 & 72.36 & 10.93 & 48.02  \\
			AcSeek~\cite{DBLP:conf/cvpr/TeneyH19} & CVPR 2019 & 46.00 & 58.24 & 29.49 & 44.33 \\
			SSL~\cite{DBLP:conf/ijcai/ZhuMLZWZ20} &   IJCAI 2020& 57.59 & 86.53 & 29.87 & 50.03  \\ 
			\hline
			CSS~\cite{DBLP:conf/cvpr/0016YXZPZ20} &   CVPR 2020& 41.16 & 43.96 & 12.78 & 47.78  \\ 
			RaReg~\cite{DBLP:conf/nips/TeneyAKSKH20} &  NeurIPS 2020& 55.37 & 83.39 & 41.60 & 44.20  \\ 
			GCE-DQ~\cite{DBLP:conf/iccv/HanWSH021} &  ICCV 2021 & 57.32 & 87.04 & 27.75 & 49.59  \\ 
			AdaVQA~\cite{DBLP:conf/ijcai/GuoNCJZB21} &   IJCAI 2021& 54.57 & 72.47 & \color{blue}{\textbf{53.81}} & 45.58  \\ 
			LRS~\cite{DBLP:journals/tip/GuoNCTZ22} &   TIP 2022& 47.09 & 68.42 & 21.71 & 42.88  \\ 
            \hline
			\textbf{UpDn+Ours} &  - & \color{blue}{\textbf{59.45}}  & 89.24 & 42.39 & \color{blue}{\textbf{50.67}}  \\ 
			\hline
		\end{tabular}
	}
	%\vspace{-9pt}
\end{table}

\subsection{Implementation Details}
In this paper, we mainly evaluated our method based on UpDn~\cite{DBLP:conf/cvpr/0004SDW0H018}, and extend our model to two backbones, namely, BAN~\cite{DBLP:conf/nips/KimJZ18} and SAN~\cite{DBLP:conf/cvpr/YangHGDS16}. Following the baseline UpDn, we used the pre-trained Faster R-CNN~\cite{DBLP:conf/nips/RenHGS15} to extract object features. For each image, it is encoded as a set of $N$=36 objects with corresponding 2048-dimensional feature vector. As to each question, the words are initialized by the 300-dimensional Glove~\cite{DBLP:conf/emnlp/PenningtonSM14} embeddings and then fed into GRU~\cite{DBLP:conf/emnlp/ChoMGBBSB14} to derive a sentence-level representation with the dimension of 1280. The hidden dimension of $\bm{\mu}_{i}$ and $\bm{\sigma}_{i}$ in the information bottleneck modulator is set to 128.

In practice, we respectively set $T_{0}$, $T_{1}$, and $T_{2}$ to 12, 14, and 20. We pre-trained the model with the VQA loss $\mathcal{L}_{vqa}$ and information bottleneck loss $\mathcal{L}_{vib}$ for $T_{0}$ epochs. Pertaining to the finetuning phrase, we first added $\mathcal{L}_{b}$ for $T_{1}-T_{0}$ epochs, and then we continued fine-tuning by adding $\mathcal{L}_{b}$, $\mathcal{L}_{c}$, and $\mathcal{L}_{b}$ for $T_{2}-T_{1}$ epochs. Besides, $\lambda_{s}$, $\lambda_{c}$, and $\lambda_{b}$ in Eq.~(\ref{eq:14}) are set to 4.0, 2.0, and 0.001, respectively. We used the Adam optimizer~\cite{DBLP:journals/corr/KingmaB14} with $\beta_{1}=0.9$ and $\beta_{2}=0.98$. The base learning rate is set to $ \mathrm{min}(2.5te^{-5}, 1e^{-4}) $, where $t$ is the current epoch number starting from 1. After 16 epochs, the learning rate is decayed by 1/4 in every two epochs to $1te^{-4}$. Our model is implemented in PyTorch framework with a NVIDIA GeForce GTX TITAN Xp GPU. The batch size $B$ is set to 64.

\begin{figure*}
  \hspace{-1.5em}
  \subfigure[Frequency statistic of right answering.]{\includegraphics[width=66mm]{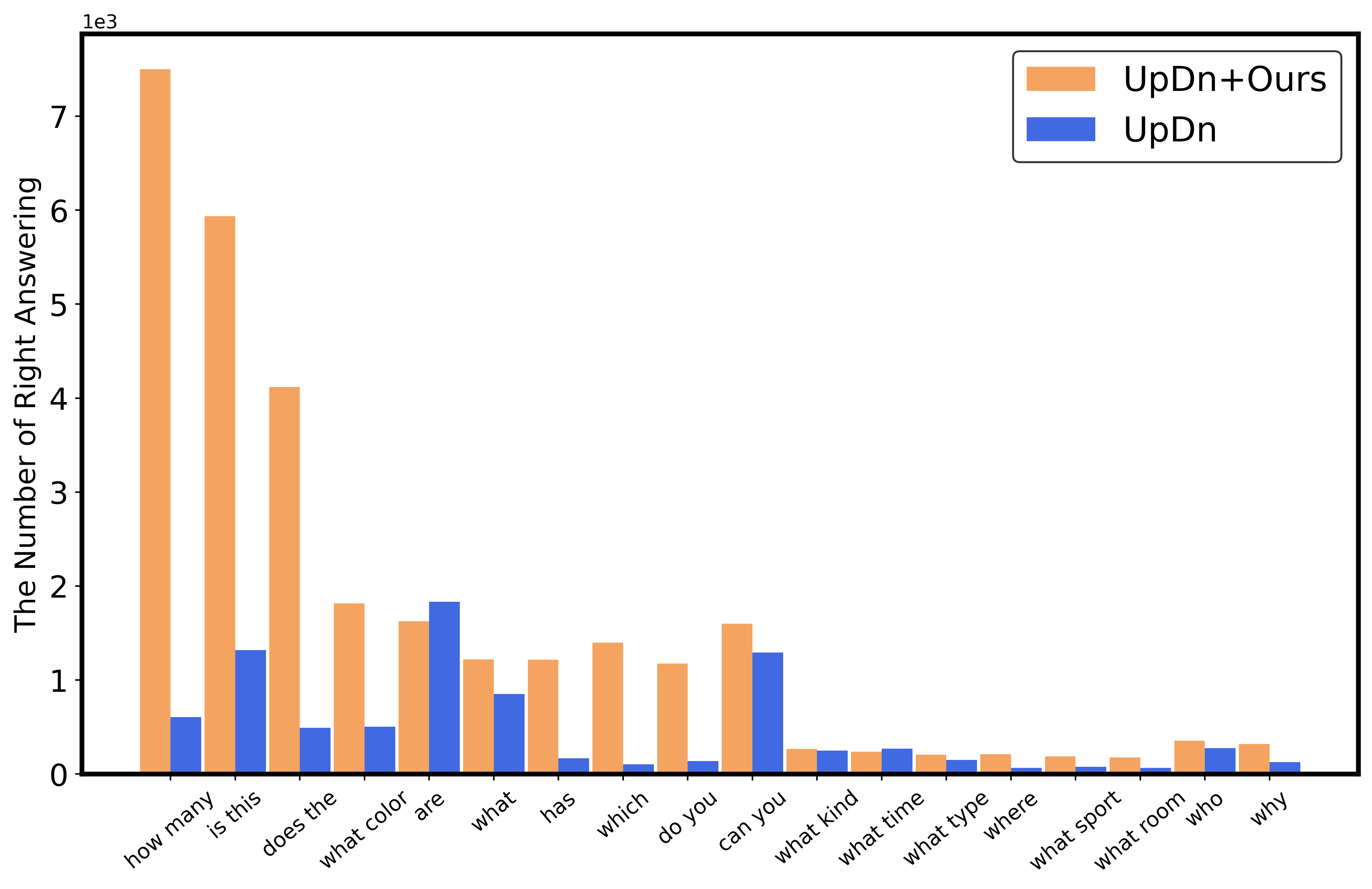}}
  \subfigure[UpDn on six question types.]{\includegraphics[width=63mm]{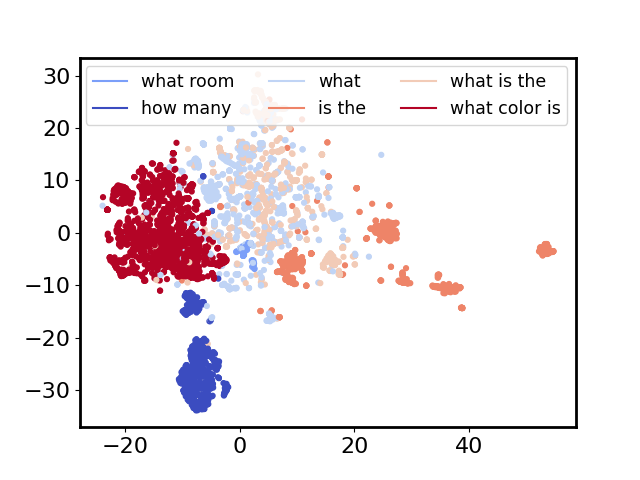}}\hspace{-1.5em}
  \subfigure[UpDn+Ours on six question types.]{\includegraphics[width=63mm]{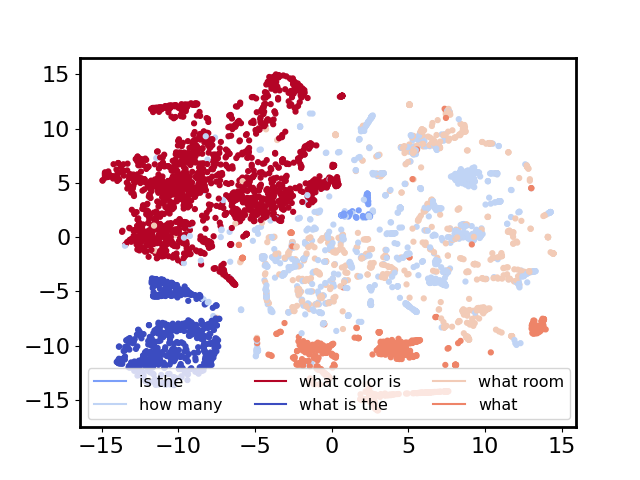}}
%   \vspace{-2em}
  \caption{(a) Frequency statistic of right answering of UpDn and UpDn+Ours on top-18 question types. (b)-(c) Visualization of the learned answer representation within different question types using T-SNE on the VQA-CP v2 \textit{test} dataset.}
  \label{fig:2}
\end{figure*}

\begin{figure}
  \hspace{-1em}
  \subfigure[]{\includegraphics[width=44mm]{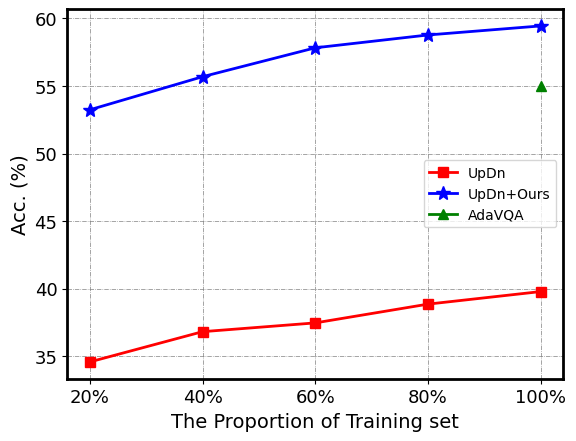}}\hspace{-0.5em}
  \subfigure[]{\includegraphics[width=45mm]{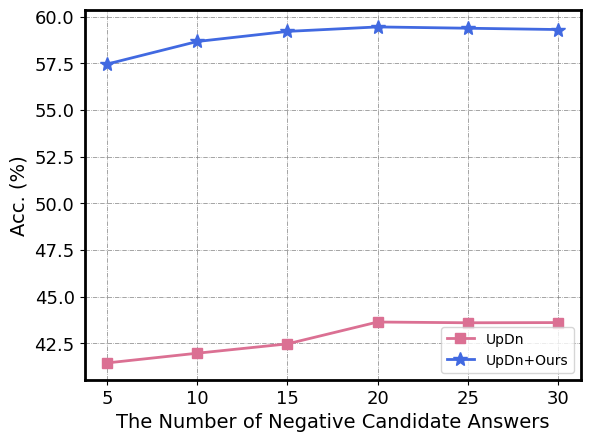}}
  \vspace{-1em}
  \caption{(a) The accuracy variation with various amounts of training dataset on UpDn and UpDn+Ours. (b) The accuracy variation with different number of the negative candidate answers.}
  \label{fig:3}
\end{figure}

\begin{figure*}
  \hspace{-1em}
  \subfigure[UpDn on \textit{what color}]{\includegraphics[width=49mm]{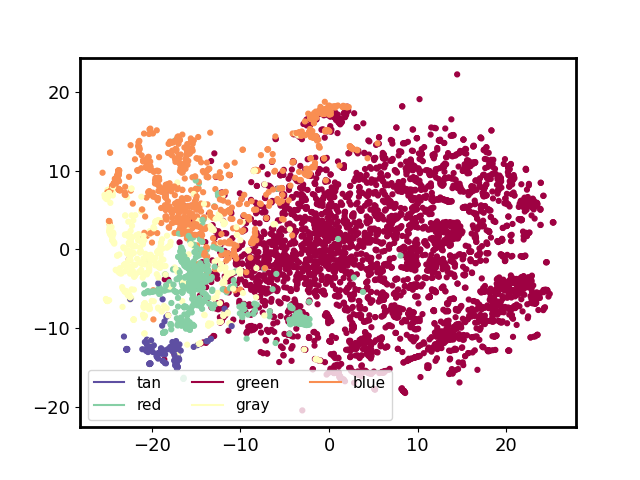}}\hspace{-1.5em}
  \subfigure[UpDn+Ours on \textit{what color}]{\includegraphics[width=49mm]{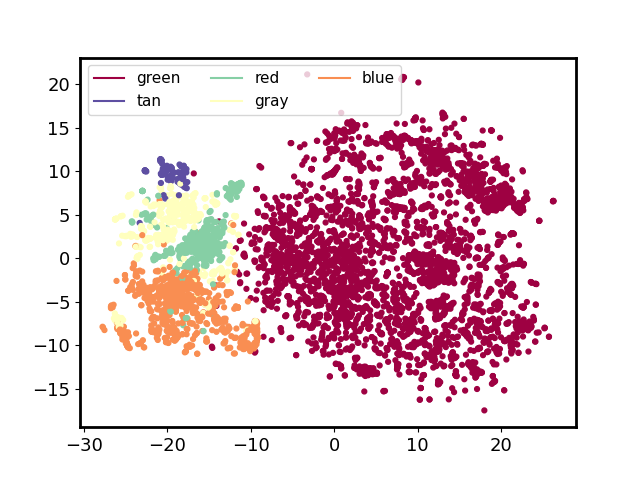}}\hspace{-1.5em}
  \subfigure[UpDn on \textit{what is}]{\includegraphics[width=49mm]{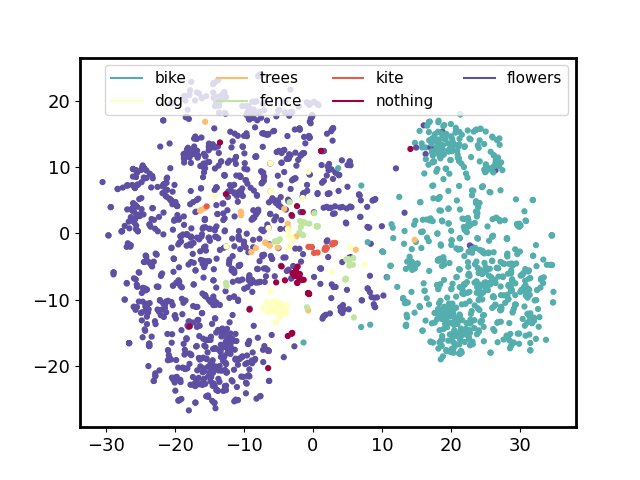}}\hspace{-1.5em}
  \subfigure[UpDn+Ours on \textit{what is}]{\includegraphics[width=49mm]{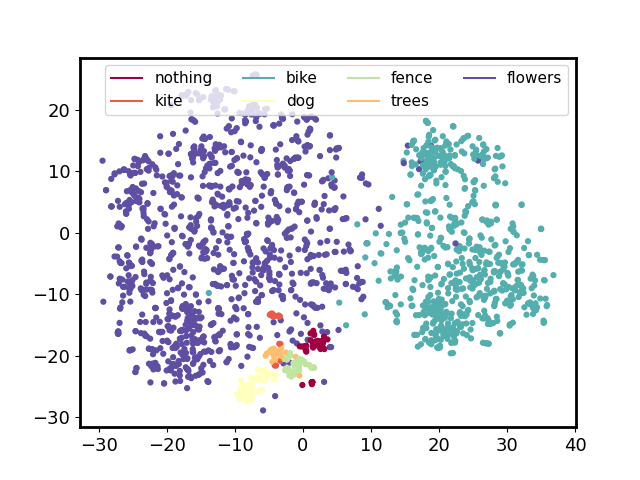}}
  \vspace{-1em}
%   \caption{(a) Performance of our LENA network with different value of scaling factor $\alpha$. (b) Number of the important regions $\Vert I(F(\mathbf{r}_{t}^{u})>0)\Vert_{0}$ at the $t$-th cell.}
  \label{fig:alpha_and_F}
\end{figure*}

\begin{figure*}
  \hspace{-1em}
  \subfigure[UpDn on \textit{how many}]{\includegraphics[width=49mm]{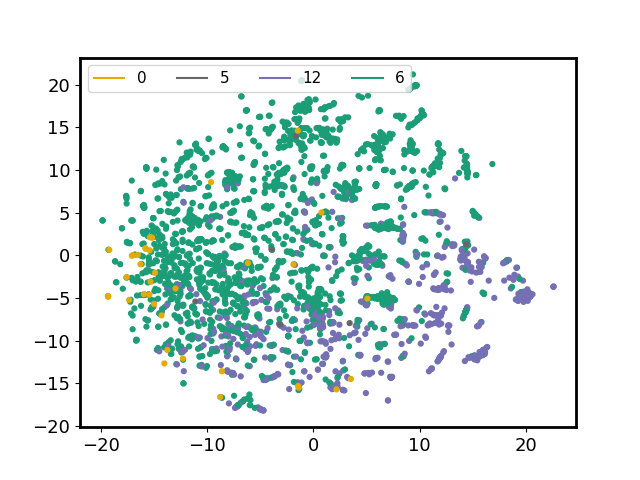}}\hspace{-1.5em}
  \subfigure[UpDn+Ours on \textit{how many}]{\includegraphics[width=49mm]{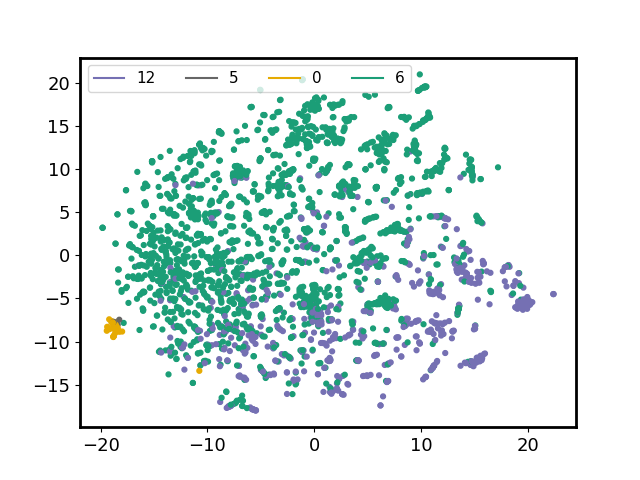}}\hspace{-1.5em}
  \subfigure[UpDn on \textit{what room}]{\includegraphics[width=49mm]{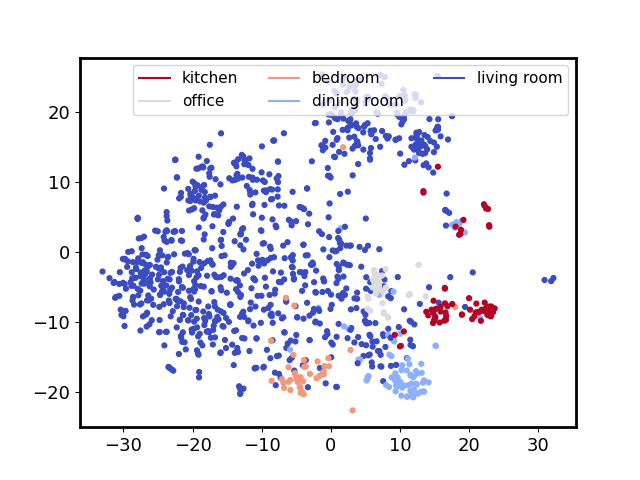}}\hspace{-1.5em}
  \subfigure[UpDn+Ours on \textit{what room}]{\includegraphics[width=49mm]{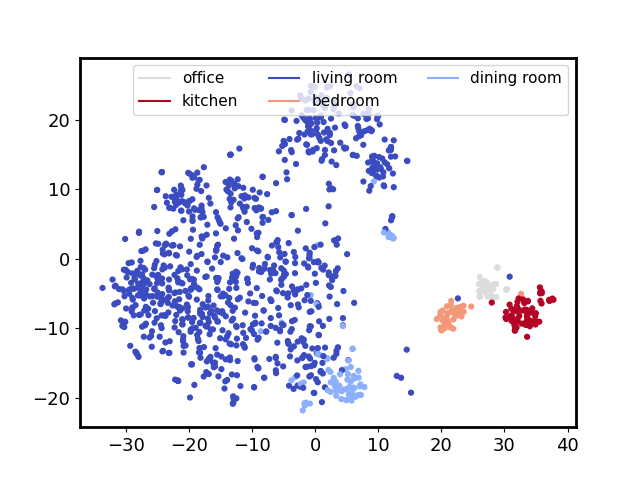}}
  \vspace{-3mm}
  \caption{(a)-(h) Visualization of the learned answer representation using T-SNE on the VQA-CP v2 \textit{test} dataset.}
  \label{fig:4}
\end{figure*}

\begin{figure}
	\includegraphics[width=88mm]{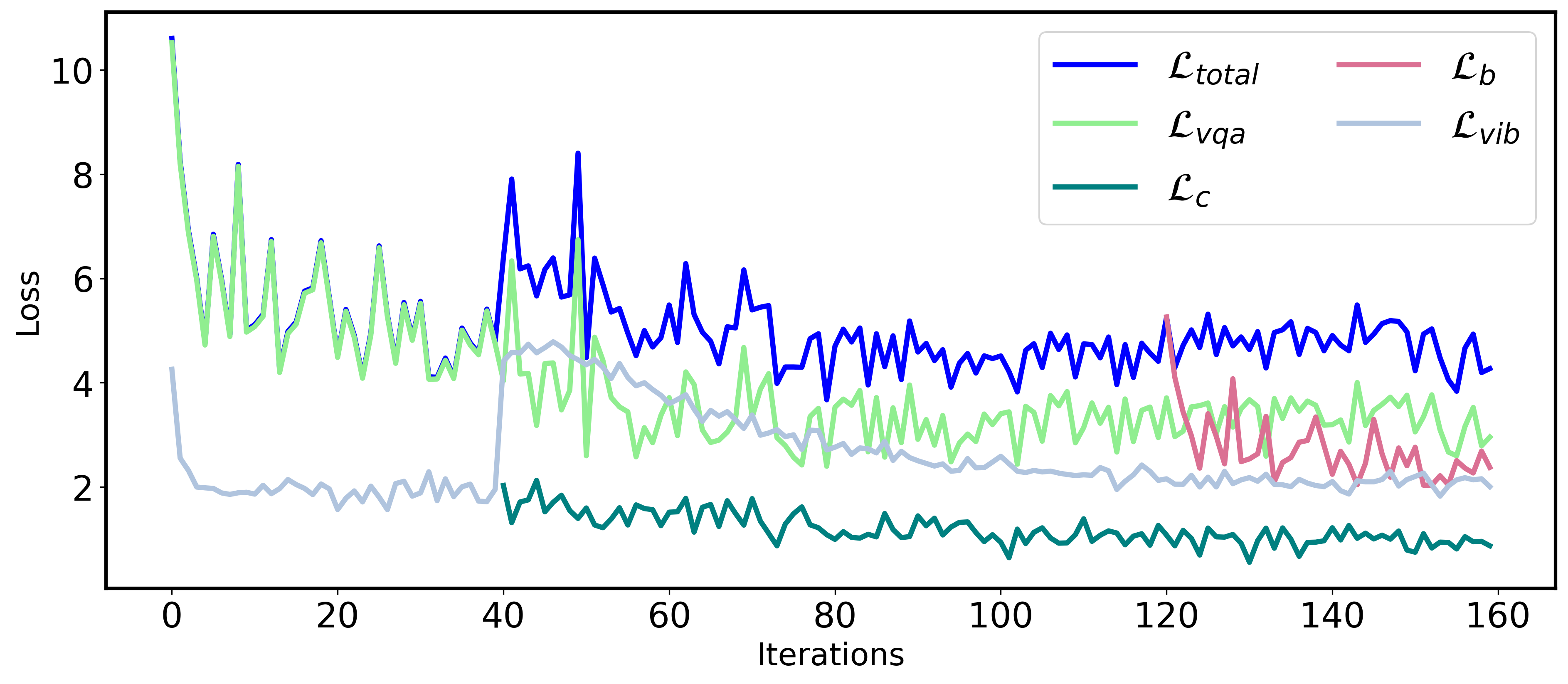}
	\vspace{-1ex}
	\caption{The convergence illustration of different losses.}
	\vspace{-3ex}
	\label{fig:5}
\end{figure}

\subsection{Performance of Accuracy Comparison (RQ1)}
Table~\ref{table:cpv1}, \ref{table:cpv2}, and \ref{table:v2} illustrate the overall and each category accuracy comparisons of our method and baselines. 
From these tables, we have the following observations:
\begin{figure*}
  \subfigure[$\lambda_{c}$=4]{\includegraphics[width=45mm]{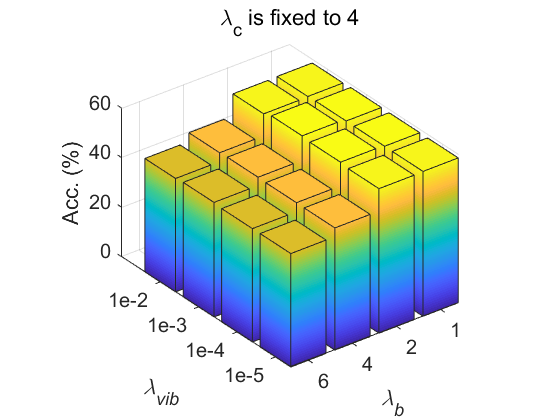}}\hspace{-0.5em}
  \subfigure[$\lambda_{b}$=2]{\includegraphics[width=45mm]{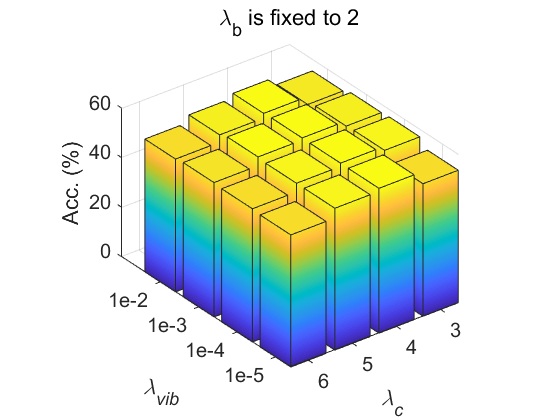}}\hspace{-0.5em}
  \subfigure[$\lambda_{vib}$=0.001]{\includegraphics[width=45mm]{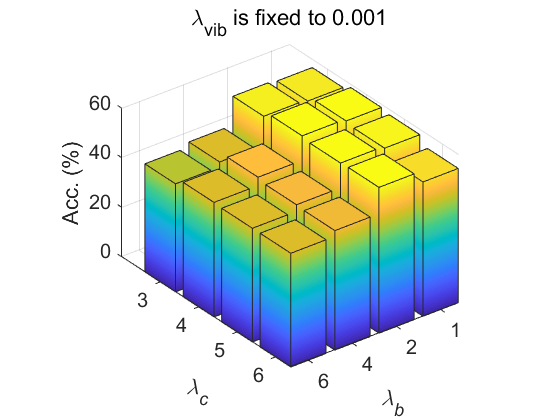}}\hspace{-0.5em}
  \subfigure[Acc. of variants w.r.t training epochs]{\includegraphics[width=48mm]{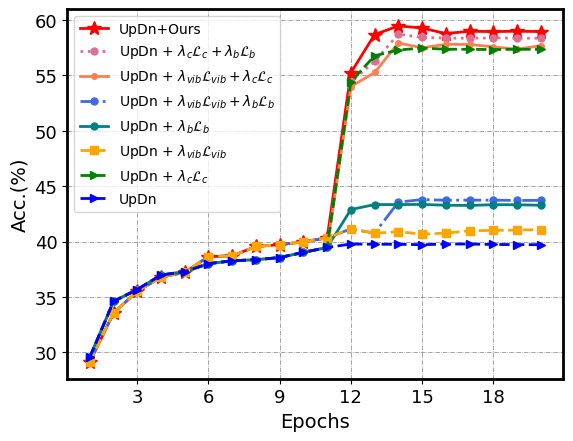}}\hspace{-0.5em}
  \vspace{-3mm}
  \caption{(a)-(c) Performance with two coefficients when fixing one. (d) Accuracy performance of different variants with respect to training epochs.}
  \label{fig:6}
\end{figure*}

\begin{table}[t]
	\caption{Performance evaluated on the $\textit{val}$ split of VQA v2.}
	\vspace{-2ex}
	\label{table:v2}
	\scalebox{1}{
		\begin{tabular}{c|c|cccc}
			\hline
			\multicolumn{1}{c|}{\multirow{2}{*}{Methods}} & \multicolumn{5}{c}{VQA v2 \textit{val} Acc.(\%)}            \\ \cline{2-6}
			\multicolumn{1}{c|}{} & Venue & All& Yes/No& Num.& Other \cr 
			\hline
			UpDn~\cite{DBLP:conf/cvpr/00010BT0GZ18} &  CVPR 2018&  63.48 & 81.18 & 42.14 & 55.66 \\
			GVQA~\cite{DBLP:conf/cvpr/00010BT0GZ18} &  CVPR 2018&  48.24 & 72.03 & 31.17 & 34.65 \\
			AdvReg~\cite{DBLP:conf/nips/RamakrishnanAL18} &   ICCV 2018 & 62.75 & 79.84 & 42.35 & 55.16 \\
			RUBi~\cite{DBLP:conf/nips/RamakrishnanAL18} &   ICCV 2018 & 61.16 & - & - & - \\
			LM~\cite{DBLP:conf/emnlp/ClarkYZ19}  &   EMNLP 2019 & 63.26 & 81.16 & 42.22 & 55.22 \\
			LMH~\cite{DBLP:conf/emnlp/ClarkYZ19}  &   EMNLP 2019 & 56.35 & 65.06 & 37.63 & 54.69 \\ 
			DLR~\cite{DBLP:conf/emnlp/FukuiPYRDR16} &   AAAI 2020 & 62.75 & - & - & -   \\ 
            VGQE~\cite{DBLP:conf/eccv/KVM20} &   ECCV 2020& \color{blue}{\textbf{64.04}} & - & - & - \\
			LPF~\cite{DBLP:conf/sigir/Liang0Z21} &   SIGIR 2021 & 62.63 & 79.51 & 42.90 & 55.02  \\
			\hline
            SCR~\cite{DBLP:conf/cvpr/AgrawalBPK18} &  CVPR 2019& 62.20 & 78.80 & 41.60 & 54.50  \\
			HINT~\cite{DBLP:conf/cvpr/AndreasRDK16} &   CVPR 2019& 63.38 & 81.18 & \color{blue}{\textbf{42.99}} & 55.56    \\ 
            SSL~\cite{DBLP:conf/ijcai/ZhuMLZWZ20} &   IJCAI 2020& 63.73 & - & - & -  \\ 
			\hline
			CSS~\cite{DBLP:conf/cvpr/0016YXZPZ20} &   CVPR 2020& 59.91 & 73.25 & 39.77 & 55.11  \\ 
 			% CF-VQA~\cite{DBLP:conf/cvpr/NiuTZL0W21} & CVPR 2021 & 63.54 & 82.51 & 43.96 & 54.30 \\
		    LRS~\cite{DBLP:journals/tip/GuoNCTZ22} &   TIP 2022& 55.50 & 64.22 & 39.61 & 53.09  \\ 
		    \hline
			\textbf{UpDn+Ours} &   - & 63.89 & \color{blue}{\textbf{81.47}}  & 41.04 & \color{blue}{\textbf{56.14}} \\
			\hline
		\end{tabular}
	}
\end{table}

% \begin{table}[t]
% 	\caption{Performance evaluated on the $\textit{val}$ split of the VQA v1.}
% 	\vspace{-2ex}
% 	\label{table:v1}
% 	\scalebox{1}{
% 		\begin{tabular}{c|c|cccc}
% 			\hline
% 			\multicolumn{1}{c|}{\multirow{2}{*}{Methods}} & \multicolumn{5}{c}{VQA v1 \textit{val} Acc.(\%)}            \\ \cline{2-6}
% 			\multicolumn{1}{c|}{} & Venue & All& Yes/No& Num.& Other& \hline
% % 			Question-Only &   ICCV' 2015& 46.75 & 77.86 & 30.24 & 27.61 \\
% 			HieCoAtt~\cite{DBLP:conf/emnlp/FukuiPYRDR16} &   CVPR 2016& 57.00 & 79.60 & 35.00 & 45.70   \\ 
% 			NMN~\cite{DBLP:conf/cvpr/AndreasRDK16} &   CVPR 2016& 54.72 & 80.44 & 34.03 & 40.66    \\ 
% 			SAN~\cite{DBLP:conf/cvpr/YangHGDS16} &   CVPR 2016& 57.60 & 78.60 & 41.80 & 46.40  \\ 
% 			\hline   
% 			UpDn~\cite{DBLP:conf/cvpr/00010BT0GZ18} &  CVPR 2018&  64.32 & 83.91 & 39.54 & 55.97 \\
%             GVQA~\cite{DBLP:conf/cvpr/AgrawalBPK18} &  CVPR 2018& 48.24 & 72.03 & - & - \\ 
%  			% SAN+Ours  & CVPR' 2016& 58.24^{\uparrow 2.38} & 81.26 & 34.61 & 50.56\\
%  			% \hline
% 			\textbf{UpDn+Ours} &   - & \color{blue}{\textbf{64.67}} & \color{blue}{\textbf{84.25}}  & \color{blue}{\textbf{39.98}} & \color{blue}{\textbf{56.30}} \\
% 			\hline
% 		\end{tabular}
% 	}
% \end{table}

\begin{figure*}
    \includegraphics[width=1\textwidth]{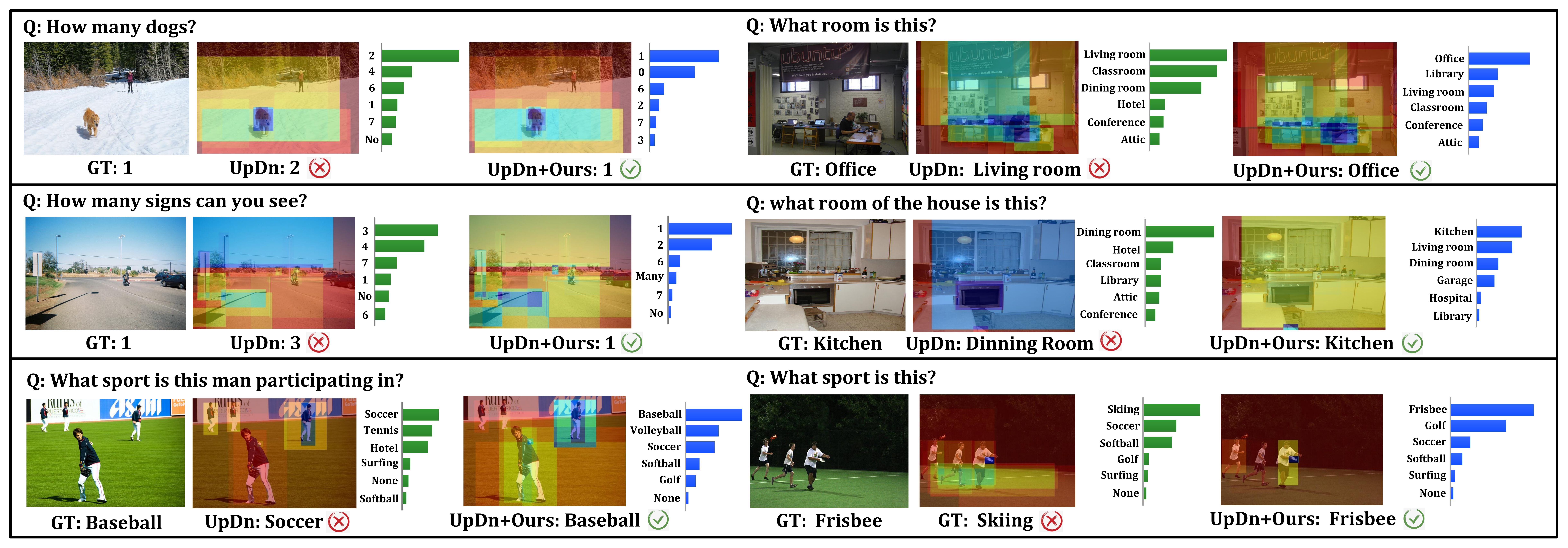}
    \vspace{-4ex}
    \caption{Qualitative comparision between our proposed model and baseline UpDn.} 
    \label{fig:attention_map}
\end{figure*}

\begin{table}[t]
	\caption{Performance on VQA-CP v2 $\textit{test}$ set based on different baselines.}
	\vspace{-3ex}
	\label{table:other_model}
	\scalebox{0.88}{
		\begin{tabular}{c|c|ccccc}
			\hline
			\multicolumn{1}{c|}{\multirow{2}{*}{Methods}} & \multicolumn{6}{c}{VQA-CP v2 \textit{test} Acc.(\%)}            \\  \cline{2-7}
			\multicolumn{1}{c|}{} & Venue & All & +$\Delta \uparrow$ & Yes/No & Num.& Other \cr 
			\hline
			SAN~\cite{DBLP:conf/cvpr/YangHGDS16} &   CVPR 2016 & 25.16 & - & 38.69 & 11.52 & 22.34  \\ 
 			SAN+Ours  & - & \color{blue}{\textbf{38.97}} & \color{blue}{\textbf{+13.81}} & \color{blue}{\textbf{50.76}} & \color{blue}{\textbf{21.74}} & \color{blue}{\textbf{28.23}} \\
 			\hline
			UpDn~\cite{DBLP:conf/cvpr/00010BT0GZ18} &  CVPR 2018 &  39.78 & - & 42.98 & 12.06 & 46.50 \\
			UpDn+Ours & - & \color{blue}{\textbf{59.45}} & \color{blue}{\textbf{+19.69}} & \color{blue}{\textbf{89.24}} & \color{blue}{\textbf{42.39}} & \color{blue}{\textbf{50.67}} \\
			\hline
			BAN~\cite{DBLP:conf/nips/KimJZ18} &   NeurlPS 2019 & 41.24 & - & 42.68 & 12.39 &  46.91 \\ 
 			BAN+Ours  & - & \color{blue}{\textbf{55.61}} & \color{blue}{\textbf{+14.37}} & \color{blue}{\textbf{76.98}} & \color{blue}{\textbf{39.68}} & \color{blue}{\textbf{48.53}} \\
 			\hline
		\end{tabular}
}
\end{table}

\begin{table}[t]
	\centering
	\caption{Effect of using different losses on VQA-CP v2.}
	  \vspace{-2ex}
	\label{table:ablation}
	\scalebox{0.83}{
		\begin{tabular}{c|c|c|c|ccccc}
			\hline
			\multicolumn{1}{c|}{\multirow{2}{*}{Method}} & \multicolumn{8}{c}{VQA-CP v2 Acc.(\%)} \\ \cline{2-9}
			\multicolumn{1}{c|}{} & $\lambda_{c}$ & $\lambda_{b}$ & $\lambda_{vib}$ & All & +$\Delta \uparrow$ & Yes/No & Num. & Other \cr
			\hline
			UpDn  & -  & - & - & 39.78 & +0.00 & 42.98 & 12.06 & 46.50 \\
 			\hline
 			UpDn+$\mathcal{L}_{c}$  & 3.0  & - & - & 53.65 & +13.01 & 80.45  & 23.42 & 47.91\\
 			UpDn+$\mathcal{L}_{c}$ & 4.0  & - & - & \color{blue}{\textbf{57.76}}  & \color{blue}{\textbf{+17.98}} & \color{blue}{\textbf{87.64}}  & \color{blue}{\textbf{40.29}} & 48.67 \\
 			UpDn+$\mathcal{L}_{c}$  & 5.0  & - & - & 57.25 & +17.47 & 85.36 & 39.47 & \color{blue}{\textbf{48.68}} \\
			UpDn+$\mathcal{L}_{c}$  & 6.0  & - & - & 56.81 & +17.03 & 84.73 & 39.13  & 48.46 \\
			 \hline
 			UpDn+$\mathcal{L}_{b}$  & -  & 1.0 & - & 42.21  & +2.43 & 47.24  & 13.17 & 47.73\\
 			UpDn+$\mathcal{L}_{b}$ & -  & 2.0 & - & 42.75  & +2.97 & 48.55  & 13.30 & \color{blue}{\textbf{47.80}}\\
 			UpDn+$\mathcal{L}_{b}$  & -  & 4.0 & - & \color{blue}{\textbf{43.65}}  & \color{blue}{\textbf{+3.87}} & \color{blue}{\textbf{50.16}} & \color{blue}{\textbf{13.86}} & 47.73 \\
 			UpDn+$\mathcal{L}_{b}$  & -  & 6.0 & - & 42.91  & +3.13 & 49.01  & 13.49 & 47.79 \\
 			\hline
 		    UpDn+$\mathcal{L}_{vib}$  & -  & - & 1e-2 & 39.67 & -0.11 & 40.15 & 12.18 & 46.23\\
 			UpDn+$\mathcal{L}_{vib}$ & -  & - & 1e-3 & \color{blue}{\textbf{41.15}} & \color{blue}{\textbf{+1.37}} & \color{blue}{\textbf{43.57}} & \color{blue}{\textbf{12.94}} & \color{blue}{\textbf{47.78}} \\
 			UpDn+$\mathcal{L}_{vib}$  & -  & - & 1e-4 & 40.60 & +0.82 & 42.42  & 12.65 & 47.31 \\
 			UpDn+$\mathcal{L}_{vib}$  & -  & - & 1e-5 & 40.33  & +0.55 & 41.81  & 12.68 & 47.15  \\
 			\hline
            UpDn+Ours  & 4.0  & 2.0  & 1e-3 & \color{blue}{\textbf{59.45}} & \color{blue}{\textbf{+19.69}} & \color{blue}{\textbf{89.24}}  & \color{blue}{\textbf{42.39}} & \color{blue}{\textbf{50.67}} \\	
			\hline
	\end{tabular}
   }
\end{table}

\begin{itemize}
	\item As shown in Table~\ref{table:cpv1} and \ref{table:cpv2}, our method achieves the highest overall accuracy, substantially surpassing all state-of-the-art baselines, on two VQA-CP benchmark datasets. These two datasets are specially developed for evaluating the capability of overcoming the language prior problem, and the comparison results demonstrate the superiority of our method on tackling the issues. Note that AdaVQA~\cite{DBLP:conf/ijcai/GuoNCJZB21} outperforms ours in terms of the \textit{Yes/No} and \textit{Num.} answer category on VQA-CP v2 and VQA-CP v1, respectively. One possible reason is that most answer types regarding \textit{Yes/No} and \textit{Num.} are less diverse, which can be easily distinguished by the marginal cosine loss proposed in AdaVQA. However, this strict constraint on the learned answer space also weakens the reasoning ability of the model to some extent. With a deeper look at the \textit{other} type (involves more ``what ...'' or ``why ...''), which has more diverse answers and requires deeper reasoning ability, we observed that our method yielded a clear-cut improvement over AdaVQA by 5.09\% and 8.68\% on VQA-CP v2 and VQA-CP v1, respectively.
	
	\item We also reported the results on the balanced VQA v2 dataset in Table~\ref{table:v2} to see whether our approach over-corrects the language bias. It is worth noting that LMH~\cite{DBLP:conf/emnlp/ClarkYZ19} yields competitive performance on VQA-CP v2 with an additional language entropy regularization. However, the accuracy drops evidently by 7.02\% on VQA v2, which indicates that the entropy penalty preoccupies the model to over-correct the language bias, especially on the question types of $\textit{Yes/No}$ and $\textit{Num.}$. By contrast, our method is more robust on VQA v2, which yields minimal performance gap between VQA v2 and VQA-CP v2.
	
	\item From the results depicted in Table~\ref{table:other_model}, we observed that the improvements for different baselines are all remarkable and consistent, which demonstrates that our method is model agnostic. In particular, as illustrated in Table~\ref{table:cpv2}, compared with other approaches with backbone model as UpDn, our method (UpDn+Ours) surpasses them by an evident margin on the VQA-CP v2 dataset.
\end{itemize}

\subsection{Component Ablation Analysis (RQ2)}
To achieve deeper insights into our proposed method, we further conducted the detailed ablation study on the VQA-CP v2 dataset. As can be seen from Table IV and Fig.~\ref{fig:6} (d), we can summarize the following conclusions:
\begin{itemize}
	\item As illustrated in Table~\ref{table:ablation}, we have recorded the accuracy performance by tuning the coefficients of different regularization losses. Particularly, when setting $\mathcal{L}_{c}$ to 4, the model can substantially outperform the base model UpDn by 17.98\%, which additionally verifies the superiority of our proposed inter-instance discriminative learning on language prior alleviation. Beyond that, the information bottleneck regularization $\mathcal{L}_{vib}$ on the latent representation space enables the model to preserve the sufficient and discriminative information, which may effectively filter the input bias. 
	\item We also depicted the accuracy variation of each variant with respect to training epochs. These variants are obtained by removing one or several regularization loss. For example, our full model surpasses the model UpDn + $\lambda_{vib} \mathcal{L}_{vib}+\lambda_{c} \mathcal{L}_{c}$ by almost 2\% point, which indicates that $\mathcal{L}_{b}$ can mitigate the language prior problem with the intra-instance invariant learning.
\end{itemize}

\subsection{Visualization of Feature Kernels (RQ3)}
We argued that language prior alleviation is attributed to discrimination of the learned answer representation space. In view of this, to obtain deeper insights into our visual perturbation-aware mechanism, we empirically visualized the learned 2-D embedding features of answers. Before that, in Fig.~\ref{fig:2} (a), we first showed the frequency statistic of right answering of UpDn and our model regarding each question types (top 18) on VQA-CP v2 \textit{test}, respectively. Clearly, as the question types that are introduced with more bias, such as $\textit{how many}$ and $\textit{what color}$, our method achieves prominent superiority over UpDn. For the question types with more diverse answers, including $\textit{what room}$, $\textit{what sport}$, and $\textit{why}$, our method also benefits from its powerful reasoning ability. Afterwards, we selected six question types to visualize their distribution of the learned answer features by UpDn and ours. As shown in Fig.~\ref{fig:2} (b) and Fig.~\ref{fig:2} (c), we observed that the learned feature points of our model on different questions types could span broader answer space. In other words, our visual-perturbation scheme can make full advantages of representation space to distinguish the answer of each instance. More importantly, the answer feature distribution within specific question type is comparatively loose. For example, the points marked with red in Fig.~\ref{fig:2} (b) can still be wide apart while the learned points in Fig.~\ref{fig:2} (c) appear in tighter pattern. This offers more possibility to further benefit the mitigation of the language prior effect. In addition, we separately visualized the internal answer representation of each question type to further illustrate the advantages of our method. Specifically, we visualized the learned features of top-$2$ answers in VQA-CP v2 $\textit{train}$ set and the answers with less frequency in VQA-CP v2 $\textit{test}$. Pertaining to the question type $\textit{what room}$, the answer $\textit{living room}$ appears most frequently in the training set. During the testing stage, some tail answers (i.e., less frequency) tend to be incorrectly classified as \textit{living room} due to the language prior problem. Taking Fig.~(\ref{fig:4}) (g) and (h) as an example, the feature points marked with yellow (\textit{bedroom}) and gray (\textit{office}) inferenced by UpDn are intertwined with the blue feature points. On the contrary, our method achieves the satisfactory separate results with sophisticated intra-instance invariant and inter-instance discriminative learning mechanism.
\vspace{-2ex}
\subsection{Parameter Analysis (RQ4)}
We carried out experiments on the VQA-CP v2 dataset to explore how the number of negative candidate answers, the various amounts of training data, and the loss coefficients (i.e., $\lambda_{c}$, $\lambda_{b}$, and $\lambda_{vib}$) affect the model performance.
\begin{itemize}
	\item To verify the advantage of our model, we conducted a series of
    experiments based on different amounts of training data randomly sampling from the original training set, and all the results are tested on VQA-CP v2. As shown in Fig.~(\ref{fig:3}) (a), we observed that our method achieves a consistent improvement over UpDn. Note that, even with 40\% training data, our approach
    can also significantly surpass the well-performing AdaVQA trained on the full training set. These observations unanimously demonstrate our approach supports great generalizability and data-insensitive reasoning capacity.
	\item We also quantitatively investigated the effect of the number of top negative candidate answers $N^{'}$. As $N^{'}$ increases, our model has potential for leveraging sufficient negative answers and intra-instance invariance learning to calibrate the feature representation, leading to an ascending performance. When $N^{'}$ is greater than 20, the performance begins to drop. This phenomenon can be explained by the fact that it is difficult for model to distinguish the candidate answers with high prediction confidence from the ground-truth, thus selecting the high confident candidates in the intra-instance invariant learning forces the model to capture more differences and facilitates the discriminative learning. However, the introduction of more candidates inevitably increases the number of low confident ones, which weaken the positive impact of high confident ones, therefore resulting in a suboptimal result.
	\item As can be seen from Fig.~(\ref{fig:6}) (a) and (c), when $\lambda_{b}$ is greater than 2, the model performance drops drastically. This is primarily due to the fact that too large value of $\lambda_{b}$ would extensively emphasize the intra-instance invariance, thereby overwhelming the fine-grained semantic discrimination. In addition, it is worth noting that both too small and too large of $\lambda_{vib}$ would deteriorate the model performance. On the one hand, the smaller $\lambda_{vib}$ can filter more input bias. One the other hand, the larger $\lambda_{vib}$ may inevitably lose some critical information for answer prediction. We also show the variation of each regularization loss with respect the iterations in Fig.~(\ref{fig:5}), to visually explain their impacts at different training stages.
\end{itemize}

\subsection{Qualitative Results (RQ5)}
\par To obtain deeper understanding about what our model learns, we presented the learned attention map and the answer prediction distribution of several instances from UpDn and our method. Taking the right bottom case in Fig.~(\ref{fig:5}) as an example, the answer \textit{soccer} appears more frequently in the training set in terms of the question type $\textit{what sport}$. This statistical bias sways the UpDn model to focus more on $\textit{grass}$ and ball (we can observe that these two objects are assigned with greater weights in the learned attention map), thereby leading to the wrong answer. On the contrary, our model leverages the visual-perturbation learning strategy to force the model to only focus on \textit{people} and \textit{frisbee} in the attention map, which provides the distilled context information for answering $\textit{frisbee}$. 

\section{Conclusion and Future Work}
\par Given a natural language question and an image, in this paper, we present a model-agnostic visual perturbation-aware framework to tackle the language prior problem in VQA. In particular, we constructed two sorts of curated images with different perturbation extents, and harnessed the combination of intra-instance invariance and inter-instance discrimination to stipulate their visual effects. Beyond that, we introduced the information bottleneck regularization and seamlessly integrated it into the learning of two discriminators, which further calibrates the latent representation for discriminative information preservation. To justify the effectiveness and scalability of our model, we conducted extensive experiments on three public benchmark datasets compared with several state-of-the-art competitors. From the experimental results, we have the following conclusions: 1) Our method achieves the clear-cut improvements over the state-of-the-art by 1.83\% in overall accuracy on VQA-CP v1 and 1.86\% on VQA CP v2. 2) Remarkably, our visual perturbation-aware learning scheme can greatly enhance the baseline models UpDn with an absolute performance gain of 19.67\% on VQA-CP v2 and 22.91\% on VQA-CP v1, strongly demonstrating its superiority of tackling the language prior problem. And 3) the visualization experiment reveals the learned representation of our model can span broader answer space and be distinguished plainly, which is more immune to the language prior. As a byproduct, we have released the data and code to facilitate researches in this community.

\par In the future, we plan to deepen and widen our work from the following two aspects: 1) We tentatively investigate the attention-perturbation learning to directly optimize the learned attention map of image for better language prior alleviation. And 2) not limited to the VQA task, we tentatively popularize our perturbation-aware learning strategy to the more general classification task and investigate its universality.


\ifCLASSOPTIONcaptionsoff
  \newpage
\fi

%\normalsize
\small
\bibliographystyle{IEEEtran}
\bibliography{tpami}

\begin{thebibliography}{10}
\providecommand{\url}[1]{#1}
\csname url@samestyle\endcsname
\providecommand{\newblock}{\relax}
\providecommand{\bibinfo}[2]{#2}
\providecommand{\BIBentrySTDinterwordspacing}{\spaceskip=0pt\relax}
\providecommand{\BIBentryALTinterwordstretchfactor}{4}
\providecommand{\BIBentryALTinterwordspacing}{\spaceskip=\fontdimen2\font plus
\BIBentryALTinterwordstretchfactor\fontdimen3\font minus
  \fontdimen4\font\relax}
\providecommand{\BIBforeignlanguage}[2]{{%
\expandafter\ifx\csname l@#1\endcsname\relax
\typeout{** WARNING: IEEEtran.bst: No hyphenation pattern has been}%
\typeout{** loaded for the language `#1'. Using the pattern for}%
\typeout{** the default language instead.}%
\else
\language=\csname l@#1\endcsname
\fi
#2}}
\providecommand{\BIBdecl}{\relax}
\BIBdecl

\bibitem{DBLP:conf/iccv/AntolALMBZP15}
S.~Antol, A.~Agrawal, J.~Lu, M.~Mitchell, D.~Batra, C.~L. Zitnick, and
  D.~Parikh, ``{VQA:} visual question answering,'' in \emph{ICCV}, 2015, pp.
  2425--2433.

\bibitem{DBLP:conf/cvpr/YangHGDS16}
Z.~Yang, X.~He, J.~Gao, L.~Deng, and A.~J. Smola, ``Stacked attention networks
  for image question answering,'' in \emph{CVPR}, 2016, pp. 21--29.

\bibitem{DBLP:conf/cvpr/00010BT0GZ18}
P.~Anderson, X.~He, C.~Buehler, D.~Teney, M.~Johnson, S.~Gould, and L.~Zhang,
  ``Bottom-up and top-down attention for image captioning and visual question
  answering,'' in \emph{CVPR}, 2018, pp. 6077--6086.

\bibitem{hu2021video}
Y.~Hu, M.~Liu, X.~Su, Z.~Gao, and L.~Nie, ``Video moment localization via deep
  cross-modal hashing,'' \emph{TIP}, vol.~30, pp. 4667--4677, 2021.

\bibitem{hu2021coarse}
Y.~Hu, L.~Nie, M.~Liu, K.~Wang, Y.~Wanga, and X.~Hua, ``Coarse-to-fine semantic
  alignment for cross-modal moment localization,'' \emph{TIP}, vol.~30, pp.
  5933--5943, 2021.

\bibitem{DBLP:conf/nips/RamakrishnanAL18}
S.~Ramakrishnan, A.~Agrawal, and S.~Lee, ``Overcoming language priors in visual
  question answering with adversarial regularization,'' in \emph{NeurIPS},
  2018, pp. 1548--1558.

\bibitem{DBLP:conf/emnlp/LiangJHZ20}
Z.~Liang, W.~Jiang, H.~Hu, and J.~Zhu, ``Learning to contrast the
  counterfactual samples for robust visual question answering,'' in
  \emph{EMNLP}, 2020, pp. 3285--3292.

\bibitem{DBLP:conf/iccv/SelvarajuLSJGHB19}
R.~R. Selvaraju, S.~Lee, Y.~Shen, H.~Jin, S.~Ghosh, L.~P. Heck, D.~Batra, and
  D.~Parikh, ``Taking a {HINT:} leveraging explanations to make vision and
  language models more grounded,'' in \emph{ICCV}, 2019, pp. 2591--2600.

\bibitem{DBLP:conf/nips/WuM19}
J.~Wu and R.~J. Mooney, ``Self-critical reasoning for robust visual question
  answering,'' in \emph{NeurIPS}, 2019, pp. 8601--8611.

\bibitem{DBLP:conf/cvpr/0016YXZPZ20}
L.~Chen, X.~Yan, J.~Xiao, H.~Zhang, S.~Pu, and Y.~Zhuang, ``Counterfactual
  samples synthesizing for robust visual question answering,'' in \emph{CVPR},
  2020, pp. 10\,797--10\,806.

\bibitem{DBLP:conf/aaai/PatroAN20}
B.~N. Patro, Anupriy, and V.~Namboodiri, ``Explanation vs attention: {A}
  two-player game to obtain attention for {VQA},'' in \emph{AAAI}, 2020, pp.
  11\,848--11\,855.

\bibitem{DBLP:conf/iccv/SelvarajuCDVPB17}
R.~R. Selvaraju, M.~Cogswell, A.~Das, R.~Vedantam, D.~Parikh, and D.~Batra,
  ``Grad-cam: Visual explanations from deep networks via gradient-based
  localization,'' in \emph{ICCV}, 2017, pp. 618--626.

\bibitem{DBLP:conf/ijcai/ZhuMLZWZ20}
X.~Zhu, Z.~Mao, C.~Liu, P.~Zhang, B.~Wang, and Y.~Zhang, ``Overcoming language
  priors with self-supervised learning for visual question answering,'' in
  \emph{IJCAI}, 2020, pp. 1083--1089.

\bibitem{DBLP:conf/cvpr/ParkHARSDR18}
D.~H. Park, L.~A. Hendricks, Z.~Akata, A.~Rohrbach, B.~Schiele, T.~Darrell, and
  M.~Rohrbach, ``Multimodal explanations: Justifying decisions and pointing to
  the evidence,'' in \emph{CVPR}, 2018, pp. 8779--8788.

\bibitem{DBLP:journals/cviu/DasAZPB17}
A.~Das, H.~Agrawal, L.~Zitnick, D.~Parikh, and D.~Batra, ``Human attention in
  visual question answering: Do humans and deep networks look at the same
  regions?'' \emph{CVIU}, vol. 163, pp. 90--100, 2017.

\bibitem{DBLP:conf/iccv/HanWSH021}
X.~Han, S.~W. andDon’t Take~the Easy Way Out: Chi~Su, Q.~Huang, and Q.~Tian,
  ``Greedy gradient ensemble for robust visual question answering,'' in
  \emph{ICCV}, 2021, pp. 1564--1573.

\bibitem{DBLP:conf/emnlp/FukuiPYRDR16}
A.~Fukui, D.~H. Park, D.~Yang, A.~Rohrbach, T.~Darrell, and M.~Rohrbach,
  ``Multimodal compact bilinear pooling for visual question answering and
  visual grounding,'' in \emph{EMNLP}.\hskip 1em plus 0.5em minus 0.4em\relax
  ACL, pp. 457--468.

\bibitem{DBLP:conf/aaai/Ben-younesCTC19}
H.~Ben{-}younes, R.~Cad{\`{e}}ne, N.~Thome, and M.~Cord, ``{BLOCK:} bilinear
  superdiagonal fusion for visual question answering and visual relationship
  detection,'' in \emph{AAAI}, 2019, pp. 8102--8109.

\bibitem{DBLP:conf/nips/KimJZ18}
J.~Kim, J.~Jun, and B.~Zhang, ``Bilinear attention networks,'' in
  \emph{NeurIPS}, 2018, pp. 1571--1581.

\bibitem{DBLP:conf/cvpr/GaoJYLHWL19}
P.~Gao, Z.~Jiang, H.~You, P.~Lu, S.~C.~H. Hoi, X.~Wang, and H.~Li, ``Dynamic
  fusion with intra- and inter-modality attention flow for visual question
  answering,'' in \emph{CVPR}, 2019, pp. 6639--6648.

\bibitem{DBLP:conf/nips/Norcliffe-Brown18}
W.~Norcliffe{-}Brown, S.~Vafeias, and S.~Parisot, ``Learning conditioned graph
  structures for interpretable visual question answering,'' in \emph{NeurIPS},
  2018, pp. 8344--8353.

\bibitem{DBLP:conf/iccv/LiGCL19}
L.~Li, Z.~Gan, Y.~Cheng, and J.~Liu, ``Relation-aware graph attention network
  for visual question answering,'' in \emph{ICCV}, 2019, pp. 10\,312--10\,321.

\bibitem{DBLP:conf/nips/WuLWD18}
C.~Wu, J.~Liu, X.~Wang, and X.~Dong, ``Chain of reasoning for visual question
  answering,'' in \emph{NeurIPS}, 2018, pp. 273--283.

\bibitem{DBLP:conf/iccv/HuARDS17}
R.~Hu, J.~Andreas, M.~Rohrbach, T.~Darrell, and K.~Saenko, ``Learning to
  reason: End-to-end module networks for visual question answering,'' in
  \emph{ICCV}, 2017, pp. 804--813.

\bibitem{DBLP:conf/cvpr/ShiZL19}
J.~Shi, H.~Zhang, and J.~Li, ``Explainable and explicit visual reasoning over
  scene graphs,'' in \emph{CVPR}, 2019, pp. 8376--8384.

\bibitem{DBLP:conf/nips/NarasimhanLS18}
M.~Narasimhan, S.~Lazebnik, and A.~G. Schwing, ``Out of the box: Reasoning with
  graph convolution nets for factual visual question answering,'' in
  \emph{NeurIPS}, 2018, pp. 2659--2670.

\bibitem{DBLP:conf/ijcai/ZhuYWS0W20}
Z.~Zhu, J.~Yu, Y.~Wang, Y.~Sun, Y.~Hu, and Q.~Wu, ``Mucko: Multi-layer
  cross-modal knowledge reasoning for fact-based visual question answering,''
  in \emph{IJCAI}, 2020, pp. 1097--1103.

\bibitem{DBLP:journals/pami/WangWSDH18}
P.~Wang, Q.~Wu, C.~Shen, A.~R. Dick, and A.~van~den Hengel, ``{FVQA:}
  fact-based visual question answering,'' \emph{TPAMI}, vol.~40, pp.
  2413--2427, 2018.

\bibitem{DBLP:conf/cvpr/MarinoRFM19}
K.~Marino, M.~Rastegari, A.~Farhadi, and R.~Mottaghi, ``{OK-VQA:} {A} visual
  question answering benchmark requiring external knowledge,'' in \emph{CVPR},
  2019, pp. 3195--3204.

\bibitem{2007DBpedia}
S.~Auer, C.~Bizer, G.~Kobilarov, J.~Lehmann, and Z.~G. Ives, ``Dbpedia: A
  nucleus for a web of open data,'' in \emph{The Semantic Web}, 2007.

\bibitem{sharma-etal-2018-conceptual}
P.~Sharma, N.~Ding, S.~Goodman, and R.~Soricut, ``Conceptual captions: A
  cleaned, hypernymed, image alt-text dataset for automatic image captioning,''
  in \emph{ACL}, 2018.

\bibitem{DBLP:conf/cvpr/AgrawalBPK18}
A.~Agrawal, D.~Batra, D.~Parikh, and A.~Kembhavi, ``Don't just assume; look and
  answer: Overcoming priors for visual question answering,'' in \emph{CVPR},
  2018, pp. 4971--4980.

\bibitem{DBLP:conf/cvpr/NiuTZL0W21}
Y.~Niu, K.~Tang, H.~Zhang, Z.~Lu, X.~Hua, and J.~Wen, ``Counterfactual {VQA:}
  {A} cause-effect look at language bias,'' in \emph{CVPR}, 2021, pp.
  12\,700--12\,710.

\bibitem{DBLP:conf/eccv/KVM20}
G.~KV and A.~Mittal, ``Reducing language biases in visual question answering
  with visually-grounded question encoder,'' in \emph{ECCV}, vol. 12358, 2020,
  pp. 18--34.

\bibitem{DBLP:conf/sigir/Liang0Z21}
Z.~Liang, H.~Hu, and J.~Zhu, ``{LPF:} {A} language-prior feedback objective
  function for de-biased visual question answering,'' in \emph{SIGIR 2021},
  2021, pp. 1955--1959.

\bibitem{DBLP:conf/nips/CadeneDBCP19}
R.~Cad{\`{e}}ne, C.~Dancette, H.~Ben{-}younes, M.~Cord, and D.~Parikh, ``Rubi:
  Reducing unimodal biases for visual question answering,'' in \emph{NeurIPS},
  2019, pp. 839--850.

\bibitem{Morgan2007Counterfactuals}
S.~L. Morgan and C.~Winship, \emph{Counterfactuals and Causal Inference}, 2015.

\bibitem{DBLP:conf/cvpr/TeneyH19}
D.~Teney and A.~van~den Hengel, ``Actively seeking and learning from live
  data,'' in \emph{CVPR}, 2019, pp. 1940--1949.

\bibitem{DBLP:conf/ijcai/GuoNCJZB21}
Y.~Guo, L.~Nie, Z.~Cheng, F.~Ji, J.~Zhang, and A.~D. Bimbo, ``Adavqa:
  Overcoming language priors with adapted margin cosine loss,'' in
  \emph{IJCAI}, 2021, pp. 708--714.

\bibitem{DBLP:journals/tip/GuoNCTZ22}
Y.~Guo, L.~Nie, Z.~Cheng, Q.~Tian, and M.~Zhang, ``Loss re-scaling {VQA:}
  revisiting the language prior problem from a class-imbalance view,''
  \emph{TIP}, vol.~31, pp. 227--238, 2022.

\bibitem{DBLP:conf/nips/RenHGS15}
S.~Ren, K.~He, R.~B. Girshick, and J.~Sun, ``Faster {R-CNN:} towards real-time
  object detection with region proposal networks,'' in \emph{NeurIPS}, 2015,
  pp. 91--99.

\bibitem{DBLP:conf/iclr/AlemiFD017}
A.~A. Alemi, I.~Fischer, J.~V. Dillon, and K.~Murphy, ``Deep variational
  information bottleneck,'' in \emph{ICLR}, 2017.

\bibitem{DBLP:conf/eccv/LinMBHPRDZ14}
T.~Lin, M.~Maire, S.~J. Belongie, J.~Hays, P.~Perona, D.~Ramanan,
  P.~Doll{\'{a}}r, and C.~L. Zitnick, ``Microsoft {COCO:} common objects in
  context,'' in \emph{ECCV}, 2014, pp. 740--755.

\bibitem{DBLP:conf/aaai/JingWZJW20}
C.~Jing, Y.~Wu, X.~Zhang, Y.~Jia, and Q.~Wu, ``Overcoming language priors in
  {VQA} via decomposed linguistic representations,'' in \emph{AAAI}, 2020, pp.
  11\,181--11\,188.

\bibitem{DBLP:conf/emnlp/ClarkYZ19}
C.~Clark, M.~Yatskar, and L.~Zettlemoyer, ``Don't take the easy way out:
  Ensemble based methods for avoiding known dataset biases,'' in \emph{EMNLP},
  2019, pp. 4067--4080.

\bibitem{DBLP:conf/nips/TeneyAKSKH20}
D.~Teney, E.~Abbasnejad, K.~Kafle, R.~Shrestha, C.~Kanan, and A.~van~den
  Hengel, ``On the value of out-of-distribution testing: An example of
  goodhart's law,'' in \emph{NeurIPS}, 2020.

\bibitem{DBLP:conf/cvpr/AndreasRDK16}
J.~Andreas, M.~Rohrbach, T.~Darrell, and D.~Klein, ``Neural module networks,''
  in \emph{CVPR}, 2016, pp. 39--48.

\bibitem{DBLP:conf/iclr/ZhangHP18}
Y.~Zhang, J.~S. Hare, and A.~Pr{\"{u}}gel{-}Bennett, ``Learning to count
  objects in natural images for visual question answering,'' in \emph{ICLR},
  2018.

\bibitem{DBLP:conf/nips/LuYBP16}
J.~Lu, J.~Yang, D.~Batra, and D.~Parikh, ``Hierarchical question-image
  co-attention for visual question answering,'' in \emph{NeurIPS}, 2016, pp.
  289--297.

\bibitem{DBLP:conf/cvpr/0004SDW0H018}
C.~Ma, C.~Shen, A.~R. Dick, Q.~Wu, P.~Wang, A.~van~den Hengel, and I.~D. Reid,
  ``Visual question answering with memory-augmented networks,'' in \emph{CVPR},
  2018, pp. 6975--6984.

\bibitem{DBLP:conf/emnlp/PenningtonSM14}
J.~Pennington, R.~Socher, and C.~D. Manning, ``Glove: Global vectors for word
  representation,'' in \emph{ACL}, 2014, pp. 1532--1543.

\bibitem{DBLP:conf/emnlp/ChoMGBBSB14}
K.~Cho, B.~van Merrienboer, {\c{C}}.~G{\"{u}}l{\c{c}}ehre, D.~Bahdanau,
  F.~Bougares, H.~Schwenk, and Y.~Bengio, ``Learning phrase representations
  using {RNN} encoder-decoder for statistical machine translation,'' in
  \emph{EMNLP}, 2014, pp. 1724--1734.

\bibitem{DBLP:journals/corr/KingmaB14}
D.~P. Kingma and J.~Ba, ``Adam: {A} method for stochastic optimization,'' in
  \emph{ICLR}, 2015.

\end{thebibliography}

\vspace{-4.0em}
\begin{IEEEbiography}	[{\includegraphics[width=1in,height=1.25in,clip,keepaspectratio]{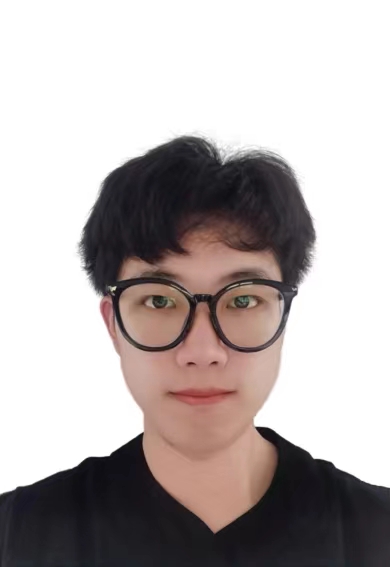}}]{Yudong Han}  received the B.E. degree from Shandong Normal University in 2020. He is currently pursuing the master’s degree with the School of Computing Science and Technology, Shandong University. His research interests include multimedia
computing and image processing. He has served as a Reviewer for various conferences and journals, such as ACM MM, AAAI, and IEEE TCYB.
\end{IEEEbiography}
\vspace{-4.3em}
\begin{IEEEbiography}
	[{\includegraphics[width=1in,height=1.25in,clip,keepaspectratio]{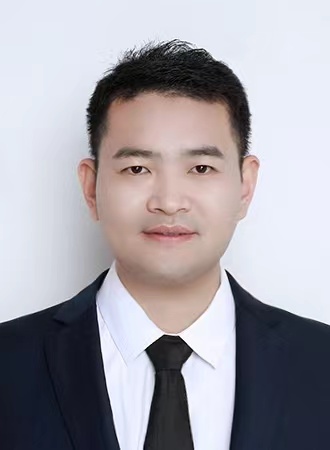}}]{Liqiang Nie} is currently the dean with the Department of Computer Science and Technology, Harbin Institute of Technology (Shenzhen). He received his B.Eng. and Ph.D. degree from Xi'an Jiaotong University and National University of Singapore (NUS), respectively. After PhD, Dr. Nie continued his research in NUS as a research fellow for three years. His research interests lie primarily in multimedia computing and information retrieval. Dr. Nie has co-/authored more than 100 papers and 4 books, received more than 14,000 Google Scholar citations. He is an AE of IEEE TKDE, IEEE TMM, IEEE TCSVT, ACM ToMM, and Information Science. Meanwhile, he is the regular area chair of ACM MM, NeurIPS, IJCAI and AAAI. He is a member of ICME steering committee. He has received many awards, like ACM MM and SIGIR best paper honorable mention in 2019, SIGMM rising star in 2020, TR35 China 2020, DAMO Academy Young Fellow in 2020, and SIGIR best student paper in 2021.
\end{IEEEbiography}
\vspace{-4.25em}
\begin{IEEEbiography}
	[{\includegraphics[width=1in,height=1.25in,clip,keepaspectratio]{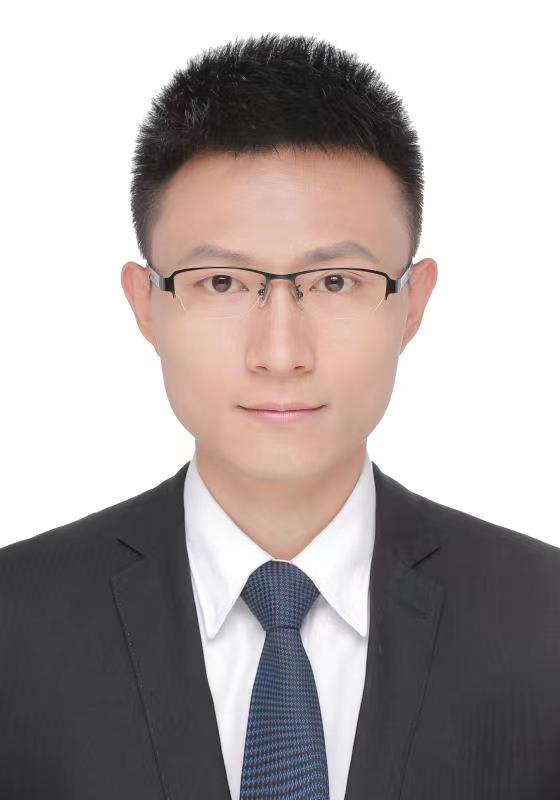}}]{Jianhua Yin} received the Ph.D. degree in computer science and technology from Tsinghua University, Beijing, China, in 2017. He is currently an Associate Professor with the School of Computer Science and Technology, Shandong University, Jinan, China. He has published several papers in the top venues, such as ACM TOIS, IEEE TMM, ACM MM, ACM SIGKDD, and ACM SIGIR. His research interests include data mining and machine learning applications.
\end{IEEEbiography}
\vspace{-4.2em}
\begin{IEEEbiography}
	[{\includegraphics[width=1in,height=1.25in,clip,keepaspectratio]{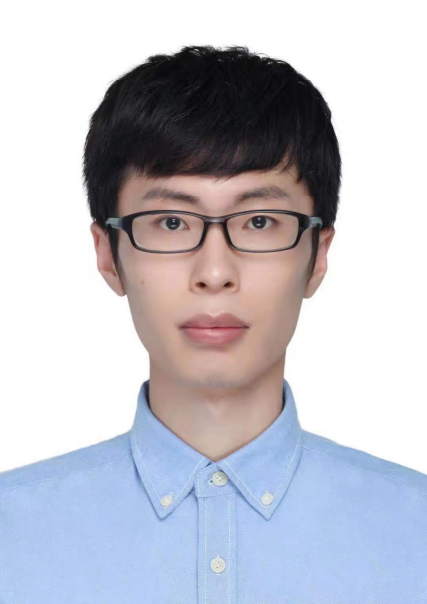}}]{Jianlong Wu} (Member, IEEE) received his B.Eng. and Ph.D. degrees from Huazhong University of Science and Technology in 2014 and Peking University in 2019, respectively. He is currently an assistant professor with the School of Computer Science and Technology, Shandong University. His research interests lie primarily in computer vision and machine learning. He has published more than 30 research papers in top journals and conferences, such as TIP, ICML, NeurIPS, and ICCV. He received many awards, such as outstanding reviewer of ICML 2020,  and the Best Student Paper of SIGIR 2021. He serves as a Senior Program Committee Member of IJCAI 2021, an area chair of ICPR 2022/2020, and a reviewer for many top journals and conferences, including TPAMI, IJCV, ICML, and ICCV.
\end{IEEEbiography}
\vspace{-4.3em}
\begin{IEEEbiography}
	[{\includegraphics[width=1in,height=1.25in,clip,keepaspectratio]{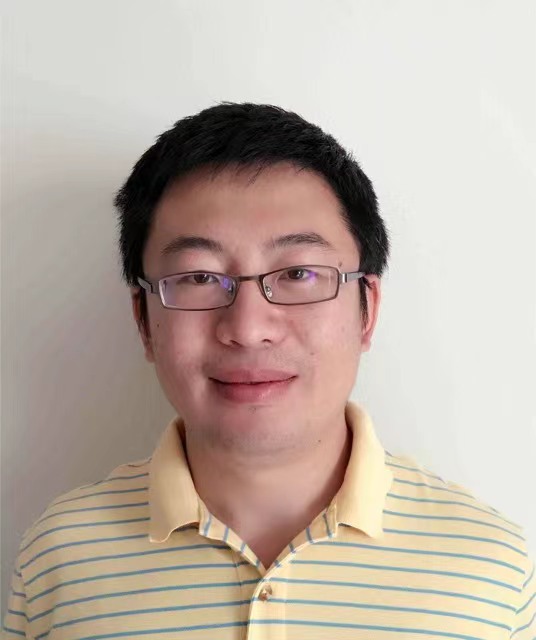}}]{Yan Yan} is currently a Gladwin Development Chair Assistant Professor in the Department of Computer Science at Illinois Institute of Technology. He was an assistant professor at the Texas State University, a research fellow at the University of Michigan and the University of Trento. He received his Ph.D. in Computer Science at the University of Trento. His research interests include computer vision, machine learning and multimedia.
\end{IEEEbiography}


%\vfill


\end{document}